%% file: main.tex
\pdfoutput=1

\documentclass[11pt]{article}

\usepackage[final]{acl}

\usepackage{times}
\usepackage{latexsym}

\usepackage[T1]{fontenc}
\usepackage[utf8]{inputenc}

\usepackage{microtype}

\usepackage{inconsolata}

\usepackage{graphicx}
\usepackage{amsmath}
\usepackage{amssymb}
\usepackage{booktabs}
\usepackage{algorithm}
\usepackage{algorithmic}

\usepackage{xcolor}
\usepackage{caption}
\usepackage{subcaption}
\usepackage{multirow}
\usepackage{makecell}
\usepackage[most]{tcolorbox} %
\usepackage{hyperref}
\usepackage{xspace}

\definecolor{bluePP}{HTML}{0070c0} %
\definecolor{yellowPP}{HTML}{ffc000} %
\definecolor{redPP}{HTML}{ff0000} %
\definecolor{greenPP}{HTML}{00b050} %

\newcommand{\highlightR}[1]{\textcolor{redPP}{#1}}
\newcommand{\highlightY}[1]{\textcolor{yellowPP}{#1}}
\newcommand{\highlightG}[1]{\textcolor{greenPP}{#1}}
\newcommand{\highlightB}[1]{\textcolor{bluePP}{#1}}

\newcommand{\executable}{\textit{executable}\xspace}
\newcommand{\partiallyexecutable}{\textit{partially-executable}\xspace}
\newcommand{\subgoalpartiallyexecutable}{\textit{subgoal-partially-executable}\xspace}
\newcommand{\nonexecutable}{\textit{non-executable}\xspace}

\tcbset{
    breakable,
    colframe=black!20,
    boxsep=5pt,
    arc=4pt,
    auto outer arc,
    fontupper=\small\ttfamily, %
    fontlower=\small\ttfamily,
    valign=top, %
    left=5pt, right=5pt, top=5pt, bottom=5pt, %
}
\title{\texorpdfstring{$How^{2}$}{How2}: How to learn from procedural How-to questions}

\author{Gautier Dagan \\
  University of Edinburgh\\
  \texttt{gautier.dagan@ed.ac.uk} \\\And
  Frank Keller \\
  University of Edinburgh\\
  \texttt{keller@inf.ed.ac.uk} \\\And
  Alex Lascarides \\
  University of Edinburgh\\
  \texttt{alex@inf.ed.ac.uk} \\ 
}

\begin{document}
\maketitle
\begin{abstract}
    An agent facing a planning problem can use answers to how-to questions to reduce uncertainty and fill knowledge gaps, helping it solve both current and future tasks.
    However, their open ended nature---valid answers to ``How do I X?'' range from executable actions to high-level descriptions of X's sub-goals---makes them challenging for AI agents to ask, and for AI experts to answer, in ways that support efficient planning.
    We introduce $How^{2}$, a memory agent framework that enables agents to ask how-to questions, store the answers, and reuse them for lifelong learning in interactive environments.
    We evaluate our approach in Plancraft, a Minecraft crafting environment, where agents must complete an assembly task by manipulating inventory items.
    Using teacher models that answer at varying levels of abstraction, from executable action sequences to high-level subgoal descriptions, we show that lifelong learning agents benefit most from answers that are abstracted and decoupled from the current state.
    $How^{2}$ offers a way for LLM-based agents to improve their planning capabilities over time by asking questions in interactive environments.
\end{abstract}

\begin{figure}[t]
    \centering
    \scalebox{0.80}{
        \includegraphics[width=\linewidth]{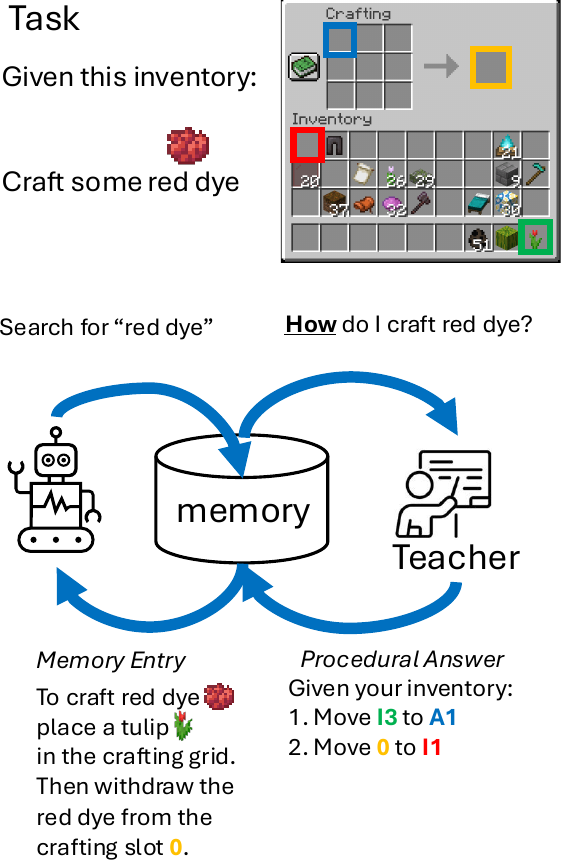}
    }
    \caption{We solve a Minecraft planning task through a lifelong mechanism in a student/teacher setup. We use a memory to store procedural answers to how-to questions. Our $How^2$ framework abstracts the executable plans, to decouple the teacher's answers from the game state and generalise memory entries for re-use.}
    \label{fig:figure1}
\end{figure}

\section{Introduction}

Asking questions is a fundamental strategy in human learning and problem solving \citep{preschoolers_questions_2010, question_asking_in_childhood}.
While AI assistants can be \textit{proactive} in questioning their users \citep{deng_survey_proactive_dialogue_2023}, these queries are often limited to seeking clarification to resolve ambiguity \citep{deng-etal-2023-prompting, Xu_Wang_Tang_Duan_Yang_Zeng_Zhou_Sun_2019, Majumder_Rao_Galley_McAuley_2021}.
But interaction with a teacher, human, or oracle is one of the ways that an automated agent can gather information to reduce its uncertainty \citep{liu2022asking}.
This is especially critical in interactive environments, where actions have consequences and resources are constrained.

In this paper, we investigate how to learn from how-to questions, which seek procedural knowledge about completing a specific task.
We define a spectrum of teacher strategies that provide varying levels of assistance, from high-level sub-goal descriptions to a fully executable sequence of actions.
We evaluate our approach in two settings, both with a wide variety of initial states: 1) the original data split, featuring \emph{low task repetition}, and 2) a new split with \emph{high task repetition}, designed to test learning on recurring goals.
We propose a memory-driven approach that translates knowledge from how-to questions into actionable abstractions for re-use in a lifelong learning paradigm.

Our contributions are: 1) the $How^2$ framework for lifelong learning from procedural questions and answers; and 2) an analysis of different teacher models with varying levels of abstraction and their effect on future LLM planning.
Our analysis reveals a trade-off between the immediate utility of an answer and its long-term reusability.
We find that while teachers providing direct, executable actions are most effective for immediate task success, answers that offer higher-level sub-goals or abstractions are more beneficial for lifelong learning.
Specifically, our memory-driven approach demonstrates how abstracting knowledge from how-to questions enables effective re-use and improves agent performance.

\begin{figure*}[t]
    \centering
    \includegraphics[width=\linewidth]{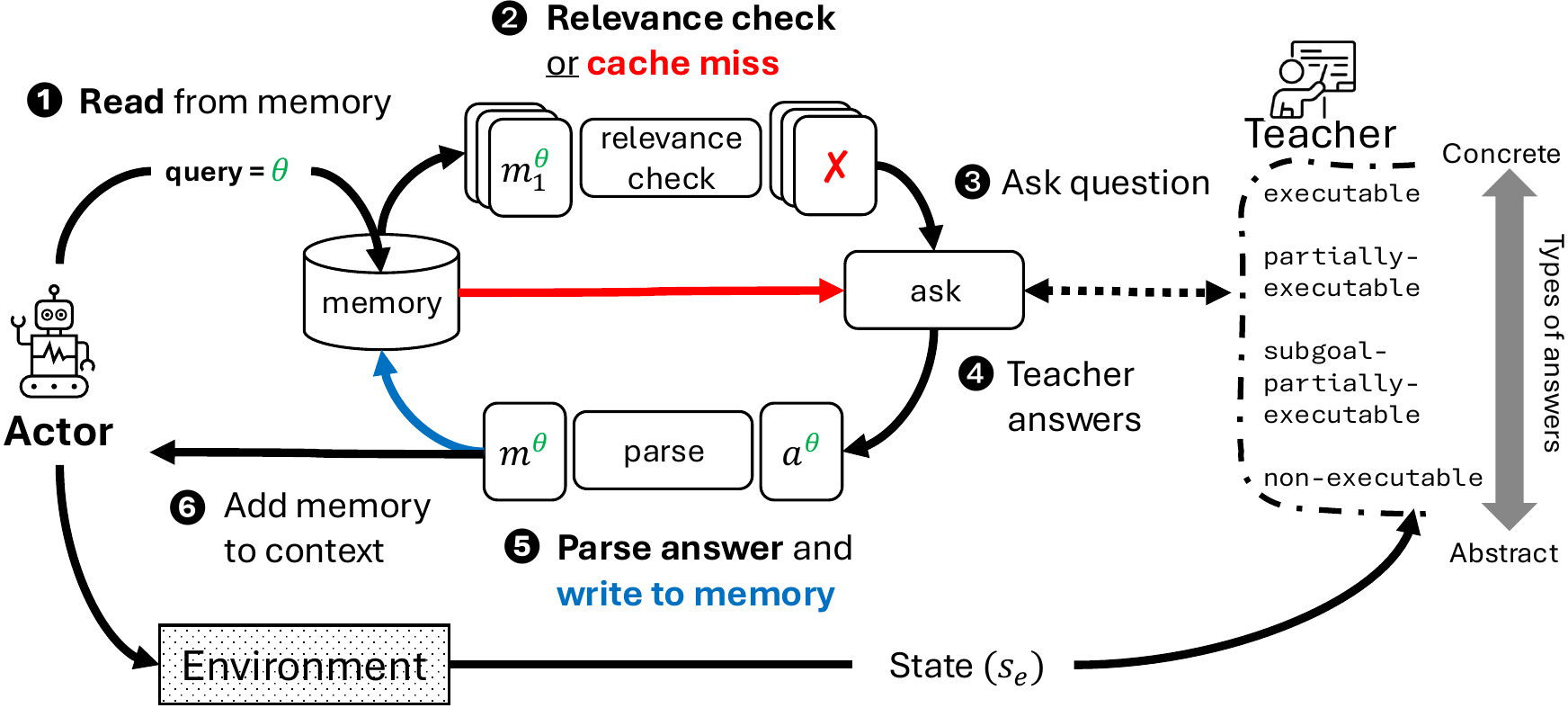}
    \caption{Our proposed $How^2$ agent framework for lifelong learning with external knowledge from a teacher. 1) The agent can call a \texttt{read-memory} tool which queries the memory module with a query $\theta$. The memory is a key-value mapping which retrieves and indexes memories given the search query $\theta$. 2) When nothing is stored under $\theta$ or all memories fail a \textbf{relevance check} w.r.t. the current state, then 3) the agent \textbf{asks} a how-to question to the teacher. 4) The teacher answers the question with different levels of executability. 5) The answer is \textbf{parsed} to decouple it from the current state and generalise the instructions. 6) The memory is stored under $\theta$ in the memory and returned to the main agent. }
    \label{fig:method}
\end{figure*}

\input{related_works.tex}

\input{method.tex}

\input{results.tex}

\section{Conclusion}

We introduced the $How^2$ framework, which enables an agent to learn from and reuse answers to procedural how-to questions.
Our experiments confirm that while fully executable plans offer the highest immediate success rate (0.59), their performance drops significantly (to 0.43) when reused, confirming \textbf{H1} that they are the least reusable.
In contrast, abstracting answers into subgoal structures enhances reusability, supporting \textbf{H2}.
The \subgoalpartiallyexecutable teacher's success rate drops by only 9\% (from 0.57 to 0.52) when answers are reused, demonstrating that abstracted knowledge generalises more effectively.

The full $How^2$ framework, which integrates memory with parsing and relevance-filtering, balances immediate utility with long-term learning.
With this framework, an agent using the \nonexecutable teacher achieves a 0.53 success rate, approaching the 0.59 performance of an agent with an always-available teacher providing fully executable plans, while reducing teacher interventions by over 40\% in a high-repetition setting (from 0.92 to 0.53).
This demonstrates that our framework enables an agent to learn procedural tasks effectively while significantly reducing its reliance on expert supervision.
Our work shows that learning from how-to questions is a powerful mechanism for improving planning capabilities, especially when answers are abstracted from the current state.

\section*{Limitations}

While our work demonstrates promising results for learning when to ask questions, several limitations should be noted.
While Plancraft enables us to compute optimal plans to simulate useful teachers—a feature absent in many public benchmarks—it remains a constrained environment compared to open-world problems where the space of possible questions is much larger.
Our parsing and relevance-checking components are tailored to the semi-structured nature of Plancraft; their generalisation to more open-ended environments would require more sophisticated natural language understanding to handle the increased variability in both the environment and potential teacher responses.
We focus exclusively on how-to questions and do not explore other question types like what-, where-, or why-questions.
Our teacher models are simulated rather than real humans, who might provide less structured or consistent responses, but might also offer richer contextual information.
We also assume that the teacher provides accurate information and do not evaluate robustness to noisy, conflicting or adversarial information.
Our memory system uses exact string matching rather than semantic search, has no forgetting mechanisms, and cannot correct stored errors.
We keep memory entries and retrieval simple to isolate how answer structure impacts learning, as exact-match retrieval avoids tuning complexities, but this can lead to memory saturation over time since incorrect memories are not updated.
Future work could address these limitations by expanding to more diverse environments, incorporating a wider range of question types, testing with human teachers, and exploring complex memory structures that rank or forget entries based on their utility.

\bibliography{custom}

\appendix

\input{appendix.tex}

\input{prompts.tex}
\input{qualitative.tex}

\end{document}

%% file: related_works.tex
\section{Related Work}

Our work sits at the intersection of three research areas: asking questions, and how best to answer and learn from them.

\paragraph{Strategic Question-Asking}
Significant previous work has focused on generating clarification questions to overcome ambiguity in dialogue and question answering \citep{Majumder_Rao_Galley_McAuley_2021, Hu_Wen_Wang_Li_de_Melo_2020, Testoni_Fernandez_2024, deng-etal-2023-prompting, White_Poesia_Hawkins_Sadigh_Goodman_2021, andukuri_stargate_2024}.
Beyond basic clarification, frameworks like Asking for Knowledge (AFK) \citep{liu2022asking}, Clarification-Execution-Planning (CEP) \citet{Zhang_Deng_Ren_Ng_Chua_2024}, and Ask-when-Needed (AwN) \citep{wang2025learning} use prompts or reinforcement learning to teach agents to query external sources or users when faced with uncertainty or unclear instructions.
Most existing work focuses on factoid questions that request missing arguments or user preferences.
We extend this research by addressing procedural how-to questions, which request sequences of actions, and by interpreting and reusing the answers in future planning problems.

\paragraph{Answering Procedural Questions}
Answering how-to questions requires presenting plan descriptions, not just facts.
Previous research has focused on formulating answers for humans by identifying structure \citep{Delpech_Saint-Dizier_2008, Saint-Dizier_2008} or tailoring retrieval for procedural content \citep{Yin_2006}.
More recently, work has explored sub-topic planning for narrative answers \citep{Cai_Yu_Liu_Yu_2022} and using graph representations to generate question-answering pairs with LLMs \citep{Pham_Hadji_Xu_Degutyte_Rainey_Kazakos_Fazly_Tzimiropoulos_Martinez_2024}.
\citet{Frummet_Elsweiler_2024} find that user preferences for answers vary with context.
Our work, in contrast, focuses on how best to present procedural information to an LLM agent.

\paragraph{Lifelong Learning from Interactions}
Our approach relates to lifelong learning, where agents improve by seeking and storing information \citep{biyik2024maple,sumers2024cognitive}.
A common method is to use a memory module to store and recall knowledge, enabling improvement without fine-tuning \citep{wang2025learning, Zhang_Deng_Ren_Ng_Chua_2024, lifelongllmsurvey, mei2025surveycontextengineeringlarge}.
For instance, systems like Retrieval-Augmented-Planning \cite{kagaya2024rap}, Reflexion \citep{reflexion}, and Memory-of-Thought \citep{li-qiu-2023-mot} accumulate past experiences or reasoning to avoid repeating mistakes.
Others focus on skill acquisition, like Voyager \citep{wang2024voyager}, which stores successful action sequences as reusable `skills'.
A third line of work investigates knowledge organisation over time, using hierarchical memory \citep{packer2024memgptllmsoperatingsystems}, knowledge networks \citep{xu2025amemagenticmemoryllm}, or structured rule libraries \citep{chen2024automanual}.
We keep memory structure simple to focus on acquiring and re-using procedural knowledge.

%% file: method.tex
\section{Method}
\label{sec:method}

We propose $How^2$, a framework for lifelong learning agents in interactive environments.
Instead of relying on trial-and-error or fine-tuning, our agent learns new multi-step procedures by asking how-to questions and reusing the answers.

\subsection{Environment}

We evaluate our agent in Plancraft \citep{dagan2024plancraftevaluationdatasetplanning}, a Minecraft crafting environment, where agents must complete an assembly task by manipulating inventory items.
This environment is well suited to test our student-teacher framework, as it contains a number of unique tasks (recipes) that all require different knowledge to solve.
Plancraft also provides a planner to benchmark against; importantly, this allows us to build a reliable Teacher agent.

Formally, let $\mathcal{E}$ be the environment (Plancraft) with observation space $\mathcal{O}$ and action space $\mathcal{A}$.
At each timestep $t$, the agent receives observation $o_t \in \mathcal{O}$ and selects action $a_t \in \mathcal{A}$.
The agent maintains a dialogue history $\mathbf{d}_t = [o_1, a_1, \ldots, o_t]$ representing the interaction sequence.
Note the observation $o_t$ is the result of the environment tool call $a_{t-1}$ executed at the previous timestep and we reserve this notation for environment observations.

\subsection{Memory}
The memory $M$ is a mutable key-value store that caches answers from the teacher.
Retrieval is based on exact string matching of the query; we do not use semantic search for simplicity.
Values are sets of memory entries associated with a query string.
We define the memory as:
\begin{align}
    M : \Theta \mapsto \mathcal{P}(\mathcal{M})
\end{align}
where $\Theta$ is the set of queries, $\mathcal{M}$ is the set of memory entries, and $\mathcal{P}(\mathcal{M})$ is the power set of memory entries.
We denote the memory entries associated with a query $\theta$ as $M[\theta]$, and each memory entry as $m^{\theta}_i \in M[\theta]$.

\subsection{\texorpdfstring{$How^2$}{How2}}

\begin{algorithm}[t]
    \begin{algorithmic}[1]
        \REQUIRE Memory $M$, Teacher $T$, observation $o_t$, query $\theta$
        \IF{$\theta \in M$}
        \STATE $relevant\_memories = \emptyset$
        \FOR{$m^{\theta}_i \in M[\theta]$}
        \IF {$\text{IsRelevant}(o_t, m^{\theta}_i)$}
        \STATE $relevant\_memories.add(m^{\theta}_i)$
        \ENDIF
        \ENDFOR
        \ENDIF
        \IF{$relevant\_memories~is~empty$}
        \STATE $q^{\theta} \leftarrow \text{AskQuestion}(o_t, \theta)$
        \STATE $r^{\theta} \leftarrow T(o_t, q^{\theta})$
        \STATE $new\_entry, \mathbf{t} \leftarrow \text{ParseAnswer}(o_t, q^{\theta}, r^{\theta})$
        \FOR {$tag \in \mathbf{t}$}
        \STATE $M[tag] \leftarrow M[tag] \cup \{new\_entry\}$
        \ENDFOR
        \RETURN $new\_entry$
        \ENDIF
        \RETURN $relevant\_memories$
    \end{algorithmic}
    \caption{$How^2$ Memory Algorithm}
    \label{alg:how2}
\end{algorithm}

$How^2$ consists of several roles: action selection, relevance check, asking and answering questions, and parsing answers (see Figure~\ref{fig:method}).
These roles can be implemented as distinct components but share information like the environment observation $o_t$.

\paragraph{Actor} The Actor is the main agent loop that determines the next action, based on the dialogue history $\mathbf{d}_t$:
\begin{align}
    \text{Actor}(\mathbf{d}_t) = a_t \in \{ & a^{\text{env}},\text{\small{\texttt{think}}(}\tau\text{\small{)}},\text{\small{\texttt{read-memory}(}}\theta\text{\small{)}}\}\nonumber
\end{align}
where $a^{\text{env}} \in \mathcal{A}$ is an environment action, $\tau$ is a thought, and $\theta$ is a query to the memory module.
All actions are expressed as tool calls, so to choose between these actions, the LLM-based Actor is required to output a tool call in the form of a valid JSON object.
An invalid tool call triggers feedback which is added to the dialogue before the Actor retries generating an action.

As in Plancraft, the Actor has access to three environment actions: \texttt{move}, \texttt{smelt}, and \texttt{impossible}.
These allow the agent to manipulate items in the inventory, smelt items in a furnace, or declare a task impossible (see Appendix~\ref{app:prompts} for tool specifications).

In this work, we add two non-environment actions: \texttt{read-memory} and \texttt{think}.
The \texttt{read-memory} action queries the memory module with a parameter $\theta$, and its implementation is shown in Algorithm~\ref{alg:how2}.
The \texttt{think} action generates a thought message $\tau$ for reasoning, similar to the \textit{think} action in Plancraft and ReAct \cite{react}.
To prevent the agent from getting stuck, we limit it to three consecutive non-environment actions before emitting a no-operation environment action.

\paragraph{Relevance Check}
When memory entries are found for a query $\theta$, we check each entry $m^{\theta}_i \in M[\theta]$ for relevance to the current game state $o_t$.
An LLM determines if the memory is applicable to the current task (see Appendix~\ref{app:prompts} for prompt).
The relevance function is defined as:
\begin{align}
    \text{IsRelevant}(o_t, m^{\theta}_i) \in \{true, false\}
\end{align}
where $m^{\theta}_i \in M[\theta]$ denotes a memory entry associated with query $\theta$.
If one or more memory entries are relevant, we append the relevant entries as a single tool response to the dialogue history.

\paragraph{Question Generation}
Otherwise, if no relevant memory entries are found, we denote this as a \textit{cache miss} (i.e., no entry exists for the current query).
For the query $\theta$ and environment observation $o_t$, we generate a how-to question~$q^{\theta}$:
\begin{align}
    \text{AskQuestion}(o_t, \theta) = q^{\theta}
\end{align}
Even though we constrain the question to a how-to question, conditioning on the observation allows the agent to refer to observed items in its questions.

\paragraph{Teacher Model}
The teacher model $T$ is a function that maps an observation and question to a procedural response:
\begin{align}
    T(o_t, q^{\theta}) = r^{\theta}
\end{align}
where $r^{\theta}$ is the teacher's response.
We explore different types of teachers conditioning and response structure (see Section~\ref{sec:teachers}).

\paragraph{Parse Answer}
Once a teacher response is obtained, we parse it and add it to memory.
The \texttt{ParseAnswer} step serves two functions.
First, it increases generalisability by abstracting state-specific details (e.g., replacing inventory slot \texttt{I12} with item name \texttt{oak\_log}).
This abstraction is key for reusing the memory entry in different states.
Second, it generates relevant tags (e.g., item names) from the answer.
We insert the parsed answer under the original query and all associated tags, enabling broader retrieval for related tasks.

ParseAnswer abstracts a teacher response for storage:
\begin{align}
    \text{ParseAnswer}(o_t, q^{\theta}, r^{\theta}) = (m^{\theta}_i, \mathbf{t})
\end{align}
where $r^{\theta}$ is the teacher's response and $q^{\theta}$ is the original question, producing a memory entry $m^{\theta}_i$ suitable for storage under query $\theta$ and a set of related tags $\mathbf{t}$.

\subsection{Teachers types}
\label{sec:teachers}

\begin{figure*}[ht!]
    \centering
    \scalebox{0.9}{
        \includegraphics[width=\linewidth]{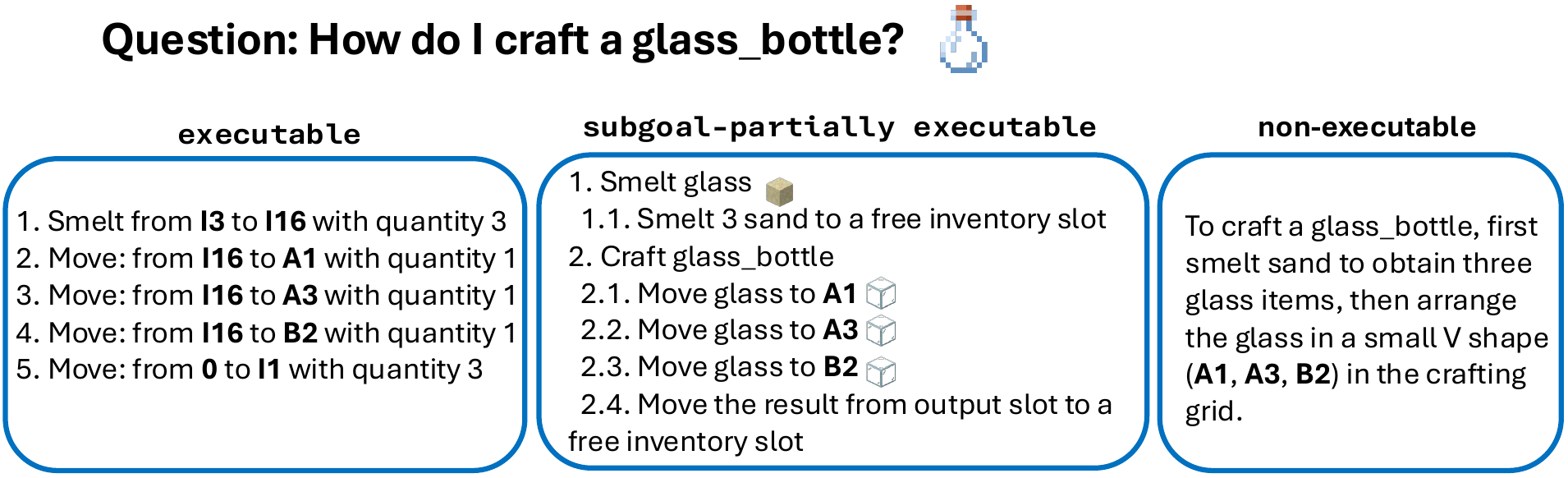}
    }
    \caption{
        The \executable teacher returns a full plan that is conditioned on the current inventory—where the inventory locations are instantiated.
        The \subgoalpartiallyexecutable teacher returns instructions where the inventory slots are not specified and decomposes each subtasks into identifiable subgoals.
        This generalises to unseen inventories as the crafting patterns remain the same.
        Lastly, the \nonexecutable teacher returns an entirely ungrounded plan and instead uses pattern abstractions such as shapes and relative positions.}
    \label{fig:teachers}
\end{figure*}

To test how different levels of abstraction in teacher responses impact an agent's learning, we design four distinct teacher models (three of which are exemplified in Figure~\ref{fig:teachers}).
These models vary in the granularity and context-dependency of the answers they provide to the agent's how-to questions.

The \executable teacher provides complete, immediately actionable plans.
It is a templated teacher that generates precise, step-by-step action sequences which can be directly execute to progress towards the goal.
These actions are fully conditioned on the current game state and thus immediately actionable.
An example answer would be `move from \texttt{I12} to \texttt{A1} with quantity 1' specifying locations and the exact parameters to pass to the `move' tool.

While executable actions are directly useful for solving the current task, they are tightly coupled to the present environment state (e.g., specific inventory slots and quantities).
This specificity makes them difficult to reuse when the underlying state changes.
We therefore hypothesise \textbf{(H1)}: \textit{executable plans are most useful for immediate execution, but the least useful for reuse in different world states}.

The \partiallyexecutable teacher is a templated teacher that offers answers that are only \textit{partially} executable.
It removes state-specific information present in the executable plan (the object positions in the inventory) and replaces them with generics that apply to all future crafting states.
For instance, instead of answering with `move from \texttt{I12} to \texttt{A1} with quantity 1', it would answer with `move the glass to \texttt{A1}'.
This is partially executable because the agent first has to identify where to retrieve the glass from in the inventory, and cannot blindly copy from the instruction.

The \subgoalpartiallyexecutable is the last templated teacher and structures the partially-executable plan into subgoals.
Instead of a list of actions, we group each set of actions into a subgoal (recipe), where each subgoal contains a list of actions.
See Figure~\ref{fig:teachers} for an example of the structure provided.
We hypothesize that \textbf{(H2)} \textit{incorporating a subgoal structure helps re-use answers and improve the effects of a memory module}.

Finally, the \nonexecutable teacher provides high-level instructions as unconstrained language (see prompt in Appendix~\ref{app:prompts}).
We expect this teacher to be the least useful as it is ungrounded in the specific naming scheme of Plancraft and instead uses pattern abstractions such as shapes and relative positions.
However, this type of teacher is closer to how a human might answer without overly relying on the specifics of the environment.
To do so, we use an LLM conditioned on planner output, observation and the agent's question.
We modify the inventory observation to abstract away the position of specific items and instead provide an aggregate view of its contents.
We also abstract away all crafting slot information (\texttt{A1},...,\texttt{C3}) and replace them with ungrounded spatial equivalents (e.g., `top left' instead of \texttt{A1}).

\subsection{Just Ask}

To evaluate our teachers and memory, we use an oracle setup (Just Ask) that bypasses the memory component.
This setup involves no lifelong learning and does not store teacher answers.
While having an expert answer every question is impractical, Just Ask serves as an upper bound for teacher performance by providing answers always tailored to the current situation.
It allows us to isolate the effectiveness of teacher responses from memory-related factors.

\subsection{Models}

We use the Llama 3.3 70B model \citep{grattafiori2024llama3herdmodels} and report further results on Qwen 3 32B \citep{yang2025qwen3technicalreport} in Appendix~\ref{app:qwen}.
As in \citet{dagan2024plancraftevaluationdatasetplanning}, we use a generation temperature of $0.6$ for the main agent role, but opt for a lower temperature of $0.2$ for all other roles.
To maintain consistency, we use the same LLM for all roles.

\subsection{Metrics}

We evaluate performance using established planning metrics: average success rate and the F1-score for correctly predicting whether a task is impossible.
Since we wish to evaluate the agent's ability to re-use knowledge, we also report the average number of cache misses per episode and the average intervention rate.
We define the average intervention rate as the ratio of the number of episodes in which the agent the teacher intervened to the total number of episodes.
A high intervention rate indicates that the agent frequently relies on asking questions, while a low rate suggests that the agent is able to solve tasks without external help and using its memory.

To evaluate $How^2$ as a lifelong learning framework, we measure the success rate as the agent is exposed to new tasks over time.
We create two dataset splits, which we refer to as \emph{low} and \emph{high} task repetition.
The \emph{low} split is the original Plancraft validation set, containing 570 examples with 347 unique tasks (targets).
The \emph{high} split is a new split we constructed from the full Plancraft dataset (train, validation, and test sets).
We select the most frequent tasks while preserving the original difficulty distribution.
This results in a dataset split with 570 examples but only 107 unique targets, meaning the agent encounters the same tasks more frequently over its lifetime (see Appendix~\ref{app:dataset-split}).

%% file: results.tex
\begin{table*}[ht]
    \centering
    \scalebox{0.9}{
        \begin{tabular}{l l cc cc cc cc}
            \toprule
                                                         &                              & \multicolumn{2}{c}{\makecell{Overall                                                                                                                 \\ SR ($\uparrow$)}} & \multicolumn{2}{c}{\makecell{Impossible\\F1 ($\uparrow$)}} & \multicolumn{2}{c}{\makecell{Avg Cache\\Miss ($\downarrow$)}} & \multicolumn{2}{c}{\makecell{Avg Intervention\\Rate ($\downarrow$)}} \\
                                                         &                              & low                                  & high                      & low                       & high                      & low  & high & low  & high \\
            \midrule
                                                         & base                         & 0.20                                 & 0.21                      & 0.43                      & 0.45                      & 0.00 & 0.00 & 0.00 & 0.00 \\
            \midrule
            \multirow{5}{*}{\rotatebox{90}{Just Ask}}    & executable                   & \textbf{\underline{0.59}}            & \underline{0.58}          & 0.93                      & 0.93                      & 1.71 & 1.60 & 0.93 & 0.92 \\
                                                         & partially-executable         & 0.54                                 & 0.53                      & 0.92                      & 0.92                      & 1.69 & 1.64 & 0.92 & 0.92 \\
                                                         & subgoal-partially-executable & 0.57                                 & 0.56                      & 0.93                      & \textbf{\underline{0.93}} & 1.56 & 1.51 & 0.92 & 0.91 \\
                                                         & non-executable               & 0.50                                 & 0.51                      & \textbf{\underline{0.94}} & 0.92                      & 1.68 & 1.56 & 0.93 & 0.92 \\
                                                         & avg                          & 0.55                                 & 0.54                      & 0.93                      & 0.92                      & 1.66 & 1.58 & 0.93 & 0.92 \\
            \midrule
            \multirow{5}{*}{\rotatebox{90}{memory-only}} & executable                   & 0.43                                 & 0.32                      & 0.74                      & 0.55                      & 0.66 & 0.28 & 0.62 & 0.26 \\
                                                         & partially-executable         & 0.48                                 & 0.41                      & 0.77                      & 0.62                      & 0.67 & 0.30 & 0.63 & 0.27 \\
                                                         & subgoal-partially-executable & \underline{0.52}                     & \underline{0.46}          & \underline{0.79}          & \underline{0.65}          & 0.66 & 0.29 & 0.63 & 0.26 \\
                                                         & non-executable               & 0.44                                 & 0.41                      & 0.78                      & 0.62                      & 0.65 & 0.29 & 0.62 & 0.27 \\
                                                         & avg                          & 0.47                                 & 0.40                      & 0.77                      & 0.61                      & 0.66 & 0.29 & 0.62 & 0.26 \\
            \midrule
            \multirow{5}{*}{\rotatebox{90}{parse}}       & executable                   & 0.48                                 & 0.44                      & 0.77                      & 0.63                      & 0.67 & 0.30 & 0.63 & 0.28 \\
                                                         & partially-executable         & 0.48                                 & 0.43                      & 0.78                      & 0.63                      & 0.67 & 0.30 & 0.63 & 0.27 \\
                                                         & subgoal-partially-executable & \underline{0.51}                     & 0.44                      & \underline{0.78}          & 0.63                      & 0.66 & 0.29 & 0.63 & 0.27 \\
                                                         & non-executable               & 0.49                                 & \underline{0.46}          & 0.77                      & \underline{0.64}          & 0.66 & 0.30 & 0.63 & 0.28 \\
                                                         & avg                          & 0.49                                 & 0.44                      & 0.78                      & 0.63                      & 0.67 & 0.30 & 0.63 & 0.28 \\
            \midrule
            \multirow{5}{*}{\rotatebox{90}{relevance}}   & executable                   & \underline{0.58}                     & \textbf{\underline{0.58}} & \underline{0.92}          & \underline{0.92}          & 1.61 & 1.58 & 0.91 & 0.89 \\
                                                         & partially-executable         & 0.52                                 & 0.50                      & 0.91                      & 0.87                      & 1.41 & 1.17 & 0.82 & 0.63 \\
                                                         & subgoal-partially-executable & 0.55                                 & 0.51                      & 0.89                      & 0.83                      & 1.03 & 0.80 & 0.80 & 0.57 \\
                                                         & non-executable               & 0.46                                 & 0.47                      & 0.88                      & 0.81                      & 0.92 & 0.68 & 0.80 & 0.56 \\
                                                         & avg                          & 0.53                                 & 0.52                      & 0.90                      & 0.86                      & 1.24 & 1.06 & 0.83 & 0.66 \\
            \midrule
            \multirow{5}{*}{\rotatebox{90}{$How^2$}}     & executable                   & 0.52                                 & 0.50                      & 0.86                      & 0.78                      & 0.94 & 0.63 & 0.79 & 0.53 \\
                                                         & partially-executable         & 0.49                                 & 0.49                      & 0.86                      & 0.77                      & 0.89 & 0.62 & 0.76 & 0.51 \\
                                                         & subgoal-partially-executable & \underline{0.53}                     & 0.50                      & \underline{0.89}          & 0.79                      & 0.87 & 0.61 & 0.78 & 0.53 \\
                                                         & non-executable               & 0.53                                 & \underline{0.53}          & 0.86                      & \underline{0.80}          & 0.83 & 0.60 & 0.77 & 0.53 \\
                                                         & avg                          & 0.52                                 & 0.50                      & 0.87                      & 0.78                      & 0.88 & 0.62 & 0.77 & 0.53 \\

            \bottomrule
        \end{tabular}
    }

    \caption{Task Success Rates, Impossible Task F1-scores, Cache Miss Rates, and Intervention Rates for the different teacher types and strategies averaged over three seeds.
        The \textbf{bold} values indicate the global best performance for each metric, while the \underline{underlined} values indicate the best performance within each group.}
    \label{tab:results}
\end{table*}

\begin{figure}[ht]
    \centering
    \scalebox{0.45}{
        \includegraphics[width=\textwidth]{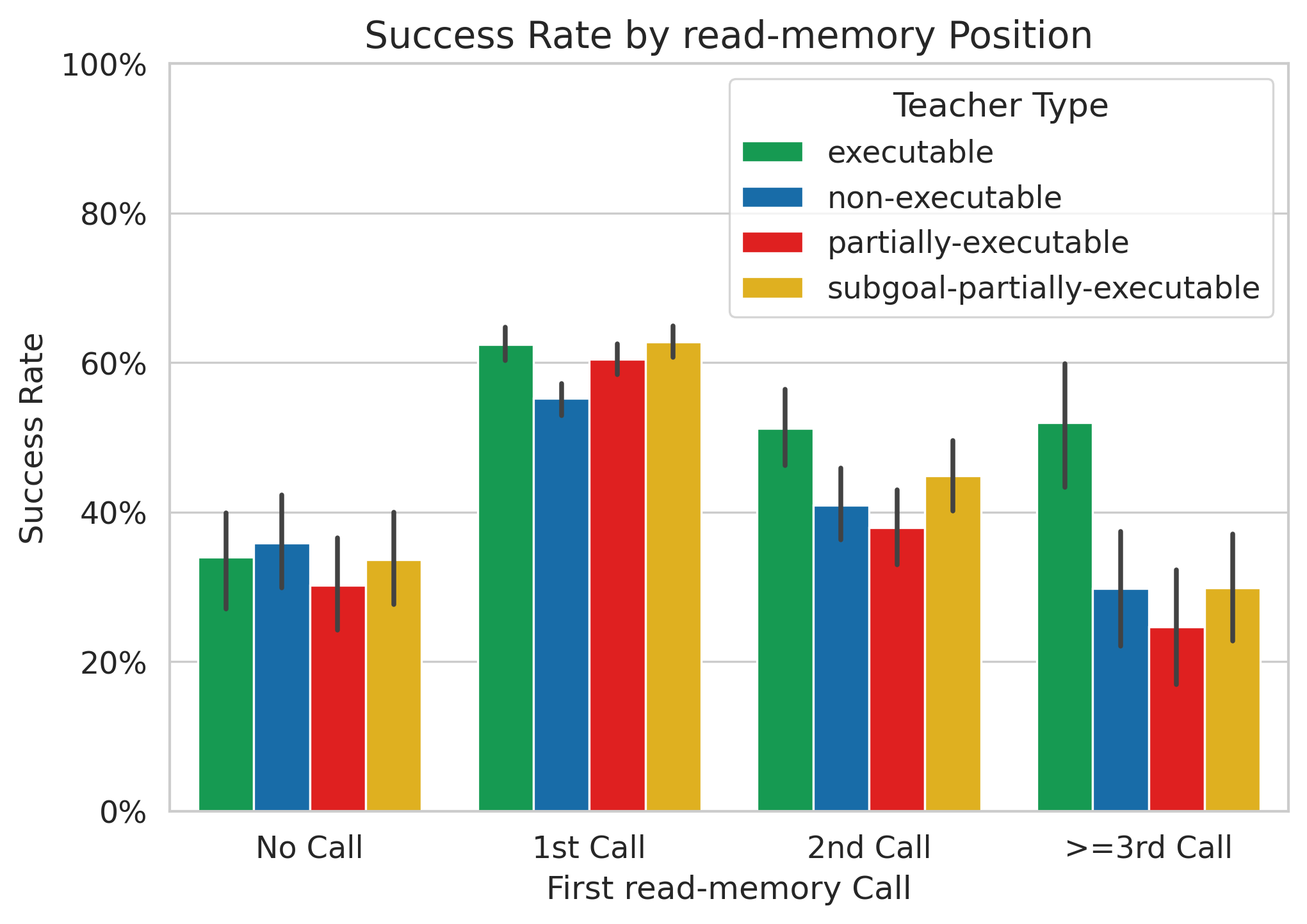}
    }
    \caption{Bar chart showing the success rate of the different teacher types in Just Ask.
        When the teacher is invoked at the beginning of the episode, the success rate is significantly higher than when it is called later.
        This is consistent across all teacher types.
        The \executable teacher outperforms all other teachers, especially if called after the first action.}
    \label{fig:results-call-position-ask}
\end{figure}

\begin{figure*}[t]
    \centering
    \includegraphics[width=\textwidth]{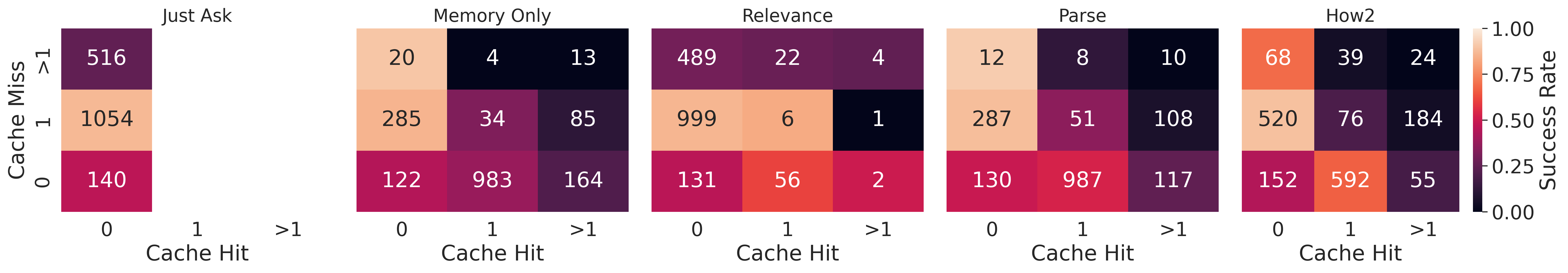}
    \caption{Heat-map for the performance of the \executable teacher in each setup. We show the success rate (colour) and counts (values) per cache misses and cache hits. This highlights the effectiveness $How^2$ in improving agent performance by filtering irrelevant memories, but also the trade-off between cache hits and success.}
    \label{fig:results-heatmap-executable-counts}
\end{figure*}

\section{Results}

We present the results of Llama 3.3 70B on both the low and high repetition settings across all teacher types and strategies in Table~\ref{tab:results}.
The base agent (no \texttt{read-memory}), without access to a teacher or memory, achieves a success rate of 0.20 (low repetition) and 0.21 (high repetition), confirming that it cannot solve Plancraft without guidance or external knowledge.
The impossible task F1-score (0.43-0.45) also indicates that it struggles to correctly identify unsolvable tasks.

\subsection{Just Ask}
We use Just Ask as a performance upper bound when the teacher answers are not stored.
All teacher types substantially improve success rates over the baseline---overall success rates range from 0.50 to 0.59.
The \executable teacher obtains 0.59 (low) and 0.58 (high) overall success, outperforming other teachers, which supports our hypothesis \textbf{H1} that executable plans are most useful for immediate execution.
However, these results are not significantly different from the \subgoalpartiallyexecutable teacher ($t=1.615$, $p=0.106$).
The structured subgoal also improves on the non-structured \partiallyexecutable ($t=1.953$, $p=0.051$) and \nonexecutable teacher ($t=4.196$, $p<0.0001$).

In Figure~\ref{fig:results-call-position-ask}, we plot the overall success rate for trajectories based on the position of the first \texttt{read-memory} call.
We find significantly higher success when the teacher is first called earlier rather than later in the episode.
This suggests that the agent benefits most from the teacher's guidance and instructions when it has not yet taken any actions.
Notably, if \texttt{read-memory} is called for the first time late in the episode (already three or more actions taken), all teachers except the \executable teacher perform worse than not consulting the teacher at all.
These results show that direct instructions from the \executable teacher are most effective when the agent is stuck or has already taken some actions.

The intervention rate (0.91-0.93) is consistently high across all teacher types, indicating that agents ask for help on nearly every task.
Finally, the high F1-scores (0.92-0.94) across all teacher types shows that agents correctly infer when task is solvable through interaction with a teacher.

\subsection{Memory}

We first introduce the memory cache without any of the other $How^2$ components (memory-only).
We find that the \executable teacher's performance drops dramatically from 0.58 to 0.32 in the high repetition split.
This degradation strongly supports the second part of \textbf{H1}: executable state-conditioned plans have low reusability (see Appendix~\ref{app:h1}).

In contrast, both  \subgoalpartiallyexecutable and \nonexecutable drop by 0.10 in the high repetition split, from 0.56 to 0.46 and from 0.51 to 0.41 respectively.
This smaller performance drop indicates that abstract answers are more generalisable, and supports \textbf{H2} that subgoal structure improves memory effectiveness.
In fact, comparing the \partiallyexecutable and \subgoalpartiallyexecutable teachers across all setups, we find that the latter performs significantly better ($t=4.050$, $p<0.0001$).

Simply using memory, the average cache miss rates (0.28-0.29) show that memory is reused in approximately $70\%$ of cases.
While, the intervention rates (0.26-0.27) demonstrate that agents now consult the teacher only about a quarter of the time.
Given that the average number of unique tasks in the high repetition split is 5.3, without accounting for re-use between target, we would expect a perfect memory to have an average intervention rate of about $0.19\ (1/5.3)$.
The lower impossible task F1-scores (0.77-0.79) are driven by lower teacher interactions and task impossibility cannot be pragmatically implied from cached answers.

\subsection{Parse}

The parse step generalises teacher answers to improve reusability.
We find that parsing slightly improves performance on average from 0.40 to 0.44 in the high repetition split.
Parsing is most effective for the \executable teacher, which improves from 0.32 to 0.44.
While only moderately effective for the \nonexecutable teacher, which improves from 0.41 to 0.46, and negative for the \subgoalpartiallyexecutable teacher, which slightly decreases from 0.46 to 0.44.
This suggests that whilst parsing increases reusability, it may only be needed if the teacher provides plans which are not already structured or generalisable (\executable) or which are entirely ungrounded (\nonexecutable).

\subsection{Relevance Check}

The relevance check adds an additional filtering step to the retrieval process, to ensure that only relevant entries are retrieved.
When used alone, it proves effective for all teachers, with an average success rate 0.52 in the high repetition split.
However, we also observe a high average intervention rate and high cache miss rates.
This is because most of the cached executable plans are irrelevant in the current state and the check enables the agent to ask the teacher directly.
Overall, the average intervention rate is 0.66 compared to the 0.92 for Just Ask, indicating that even with the relevance check which allow bypassing memory, the agent is still re-using answers.

\subsection{\texorpdfstring{$How^2$}{How2}}

The full $How^2$ framework, which integrates memory with both parsing and relevance checks, exemplifies the trade-off between performance and long-term autonomy.
While its average success rate of 0.52 (high repetition) is slightly below the \textit{Just Ask} oracle (0.54), it achieves this with a \textbf{42\% lower intervention rate} (0.53 vs. 0.92), demonstrating a significant gain in answer re-use.
The $How^2$ framework proves particularly adept at operationalising abstract knowledge.
The \nonexecutable teacher, for instance, achieves its highest success rate (0.53) within this setup.
This result highlights how the parse and relevance modules work in tandem to ground high-level, human-like instructions into reusable, actionable knowledge.

Figure~\ref{fig:results-heatmap-executable-counts} illustrates this dynamic in the \executable teacher (see Appendix~\ref{app:cache-misses} for all teachers).
Just Ask (left) yields a high success rate of 0.87 with a single query, but this plummets to 0.25 when the agent asks for help multiple times, confirming that repeated queries signal tasks beyond the agent's ability.
In contrast, the full $How^2$ framework (right) leverages its components to improve knowledge reuse.
Parsing increases the number of cache hits, while relevance checks filter out inapplicable memories.
This results in a \textbf{$67\%$ success rate from a single, relevant cache hit}, showcasing the framework's ability to make the agent a more effective and self-sufficient learner.

Finally, we also find that the $How^2$ framework is effective for reasoning models such as Qwen 3 32B (see Appendix~\ref{app:qwen}).
When using Qwen 3, $How^2$ matches the performance of the \textit{Just Ask} oracle while reducing the intervention rate by 39\%.
While overall, Qwen 3 model uses the \texttt{read-memory} action less frequently, it benefits more from the interaction with the teacher.
Surprisingly, the \nonexecutable teacher achieves the highest success rate with Qwen 3, which suggests that the reasoning capabilities improve the quality of teacher responses.

%% file: appendix.tex
\section{Repeated Dataset Split}
\label{app:dataset-split}

\begin{figure}
    \centering
    \includegraphics[width=\linewidth]{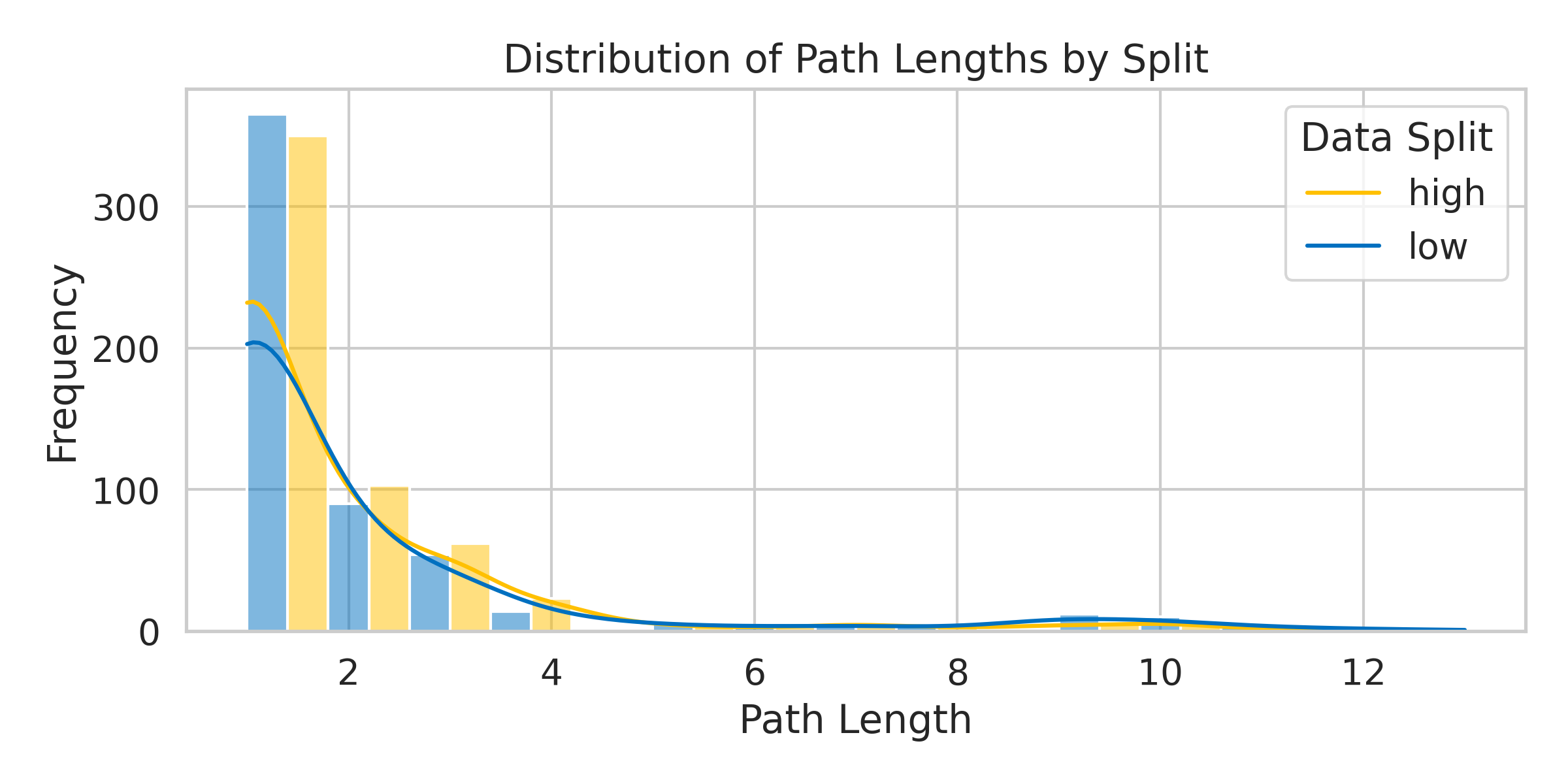}
    \caption{Distribution of path lengths for optimal plans in the low and high dataset splits. Note these path lengths represent the number of recipes needed to reach the target item and not the number of steps. There can be many steps for a single recipe. All Plancraft examples have a maximum of 30 environment steps.}
    \label{fig:path-lengths-distribution}
\end{figure}

In order to test the effectiveness of our approach in a lifelong learning setting, we create a new dataset split that contains a higher number of repeated tasks/targets.
To construct this split, we take all the examples from the original Plancraft dataset (train, validation, and test) and sort them by the most frequent tasks.
We then select a set of 570 examples from the most common tasks that follow the same complexity distribution as the original validation split.
As we show in Table \ref{tab:dataset-split-complexity}, the new split (high) has the same number of examples for each complexity level as the original validation split (low).
In Table \ref{tab:dataset-split-num-distractors}, we show that the number of distractors in the new split is similar to the original validation split.
And in Table \ref{tab:dataset-split-avg-items}, we show the average number of items used in the target plan, unique items used in the target plan, and complexity of the target plan for both splits.
Finally, we show the distribution of path lengths for optimal plans in both splits is similar (Figure \ref{fig:path-lengths-distribution}).

We call this new split the \textit{high} split as it contains a higher number of repeated tasks, and the original validation split the \textit{low} split.
Out of 570 examples, there were 347 unique tasks in the original validation split, while there are only 107 unique tasks in the new split.
This means that there is an average of 5.3 examples per task in the new split, while there is only an average of 1.6 examples per task in the original validation split.
We would therefore expect the agent to be able to re-use knowledge from previous tasks in the new split, as it has more opportunities to see the same task multiple times.
\textbf{Note that while targets/tasks are repeated, each example is unique in terms of its initial state (inventory items) and there are often multiple valid recipes for the same target.}

\begin{table}[h]
    \centering
    \begin{tabular}{lrrrr}
        \toprule
        Split & easy & medium & hard & impossible \\
        \midrule
        low   & 200  & 100    & 170  & 100        \\
        high  & 200  & 100    & 170  & 100        \\
        \bottomrule
    \end{tabular}
    \caption{Distribution of complexity for the original validation split of the Plancraft dataset (low) and the new split with additional repeated examples (high).
    }
    \label{tab:dataset-split-complexity}
\end{table}

\begin{table}[h]
    \centering
    \begin{tabular}{lrrr}
        \toprule
        Split & 4   & 8   & 16  \\
        \midrule
        low   & 182 & 206 & 182 \\
        high  & 178 & 204 & 188 \\
        \bottomrule
    \end{tabular}
    \caption{Distribution of the number of distractors in the original validation split of the Plancraft dataset (low) and the new split with additional repeated examples (high).
        The number of distractors is the number of items in the inventory that are not part of target plan.}
    \label{tab:dataset-split-num-distractors}
\end{table}

\begin{table}[h]
    \centering
    \begin{tabular}{lccc}
        \toprule
        Split & \makecell{Avg. \#                \\Items\\Used} & \makecell{Avg. \#                 \\Unique Items\\Used} & \makecell{Avg. \\Complexity} \\
        \midrule
        low   & 6.73              & 2.92 & 20.45 \\
        high  & 6.96              & 3.10 & 24.39 \\
        \bottomrule
    \end{tabular}
    \caption{Average number of items used in the target plan, unique items used in the target plan, and complexity of the target plan for the original validation split of the Plancraft dataset (low) and the new split with additional repeated examples (high).
        The complexity calculation is taken directly from the Plancraft dataset, which is based on the number of items in the target plan, number of unique items, and plan length.}
    \label{tab:dataset-split-avg-items}
\end{table}

\section{Computational Resources}
\label{app:resources}

We use vLLM \citep{vllm} to serve both the Qwen 3 32B and Llama 3.3 70B models for inference.
For all experiments, we use a node with four NVIDIA A100 GPUs with 80GB of memory.
We estimate that it would take between 600-800 GPU hours to reproduce all results.

\section{Ablations}
\label{app:ablations}

\subsection{Validating Teacher models}
\label{app:validating-teacher-models}
As mentioned in Section~\ref{sec:teachers}, we use three hard-coded teacher models (\executable, \partiallyexecutable, and \subgoalpartiallyexecutable) and one LLM-based teacher models (\nonexecutable).
We experimented with using entirely LLM-based teachers but found that the LLM-based teachers were consistently introducing errors or new-information in the plan.
In particular, there are certain recipes which the LLama 70B model almost always gets wrong, such as the \texttt{cookie} recipe.

We tested using an LLM for all teacher models on the low repetition split of the Plancraft dataset, which contains 570 examples, and use an LLM-as-a-Judge strategy to compare the teacher's answer to the expected templated planner answer.
Table~\ref{tab:teacher-models-validation} shows the results of the LLM-as-a-Judge validation on the different teachers.
We prompted the LLM to judge the teacher answer for consistency against a templated teacher and output VALID if the teacher answer is consistent and INVALID otherwise.
We found that the \nonexecutable teacher introduced the least new information or deviated the least from the planner output (after it had been parsed of state specific information).
As a result, to evaluate the effects of the teacher's answer structure in lifelong learning, we opted to use templated teachers apart from the \nonexecutable teacher.
\begin{table}[h]
    \centering

    \begin{tabular}{lc}
        \toprule
        teacher                      & \makecell{LLM Judge \\Validation ($\%$)} \\
        \midrule
        executable                   & 0.82                \\
        partially-executable         & 0.67                \\
        subgoal-partially-executable & 0.93                \\
        non-executable               & 0.95                \\
        \bottomrule
    \end{tabular}
    \caption{Validation of teacher models on the low split of the Plancraft dataset. The validator result is the percentage of answers that match the expected planner answer.}
    \label{tab:teacher-models-validation}
\end{table}

\begin{table}[h]
    \centering
    \scalebox{0.78}{
        \begin{tabular}{l cc}
            \toprule
                              & \makecell{Success Rate                 \\ ($\uparrow$)} & \makecell{Avg. Intervention \\ Rate ($\downarrow$)} \\
            \midrule
            $How^2$           & 0.50                   & \textbf{0.53} \\
            -curriculum       & 0.49                   & 0.54          \\
            +ask-first-policy & \textbf{0.52}          & 0.55          \\
            \bottomrule
        \end{tabular}}
    \caption{Comparing the average over all teachers of removing curriculum learning or using an fixed question policy over the high repetition split.
        The overall success rate is the percentage of tasks that the agent was able to complete successfully, and the average intervention rate is the average number of times the agent had to ask for help from the teacher.}
    \label{tab:ablations}
\end{table}

\subsection{Curriculum Learning}
\label{app:curriculum-learning}

As in Voyager \citep{wang2024voyager}, we test using a curriculum learning approach, where the order of the examples is sorted from recipes that have no dependencies to recipes that have dependencies (i.e. recipes that require other recipes to be completed first).
Using the all recipes and their required dependencies, we create a directed acyclic graph (DAG) where each node is a recipe and each edge is a dependency.
We then sort the recipes in the target plan by their dependencies, where recipes with no dependencies are sorted first, followed by recipes that have dependencies on other recipes.
Note that there are multiple valid orderings due to cycles in the DAG, but we remove edges belonging to cycles randomly to create a valid ordering.
Each random seed tested leads us to test a different ordering of the curriculum.

Our results are shown in Table~\ref{tab:ablations}.
Overall, we find that curriculum learning only slightly improves the performance of our agent or framework, but not significantly ($t=1.01$, $p=0.312$).
This is likely because, unlike other lifelong learning approaches that depend on experience, our student-teacher setup is less dependent on the order of examples.
If the agent is stuck or unable to solve a task, it can always ask the teacher for guidance regardless of the example order.
This knowledge still accumulates overtime but the order in which it is gathered is not as important as in other lifelong learning approaches.
Future work might explore the cost of answering from a complexity budget to see if the order of examples matters more when the teacher is able to answer a limited number of questions or only easy questions.

\subsection{Fixed Ask Policy}
\label{app:fixed-ask-policy}

We test a fixed ask policy, where the agent is required to ask a how-to question at the first turn of each new task.
This ensures that the agent always seeks clarification and guidance from the teacher before attempting to complete the task.
Forcing the agent to always ask a question first seems to slightly increase overall success at the cost of a higher intervention rate.
However these results are also not statistically significant for the tested Llama 3.3 70B model ($t=1.391$, $p=0.164$).
Since the LLama 3.3 70B model uses the \texttt{read-memory} action around $90\%$ of the time, the effects of forcing the model to ask a question first has little effect.
However, other models, such as Qwen 3 32B (see Appendix~\ref{app:qwen}), which have a lower utilisation rate of external actions might benefit from a fixed ask policy or policies that  force interaction with the memory module.

\section{Additional Results}
\label{app:additional-results}

In this section, we present additional results on Llama 3.3 70B that support our hypotheses and provide further insights into the performance of our models.

\subsection{Hypotheses Testing}
\label{app:h1}

We test our first hypothesis (\textbf{H1}) that fully executable plans are the most immediately useful but the least reusable.
To test the first part of \textbf{H1}, we compare the success rates of the Just Ask for \executable against \subgoalpartiallyexecutable.
We pick \subgoalpartiallyexecutable as it is most teacher model with the highest success rate.
The result of the t-test is inconclusive, with a p-value of 0.106, indicating that there is no significant difference between the two models in terms of success rate.
We therefore cannot validate the first part of \textbf{H1}.

Table \ref{tab:anova} presents the full results of the two-way Analysis of Variance (ANOVA).
We evaluate the effect of the plan type and the introduction of memory on the agent's success rate.
Comparing the success rates of Just Ask and mem agents for \executable and \subgoalpartiallyexecutable.
The significant interaction effect (C(teacher):C(memory)) provides statistical support that there is a relationship between teacher type and memory.
This means the effect of memory on success rate differs depending on the teacher (or vice versa).

\begin{figure}
    \centering
    \includegraphics[width=\linewidth]{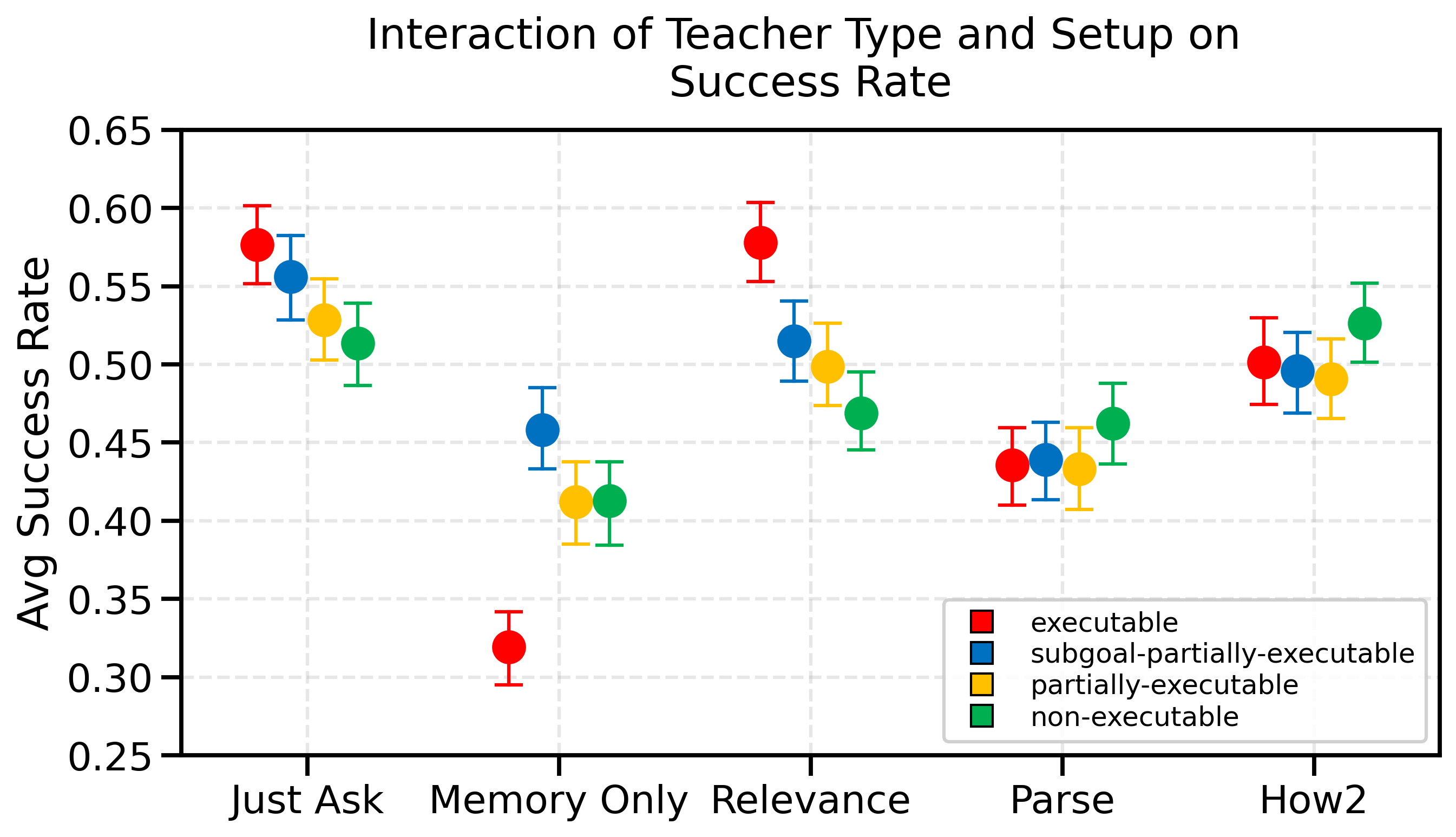}
    \caption{We plot the success rate of the different teachers for all five different memory setups.
        The \executable teacher is the least reusable, with a success rate of 0.59 when used in Just Ask, but only 0.43 when used in mem.
        The \subgoalpartiallyexecutable teacher is the most reusable, with a success rate of 0.57 when used in Just Ask, and 0.52 when used in mem.
        We see the effects of parsing teacher answers in the way that the average success rates converge.}
    \label{fig:h1_success}
\end{figure}

\begin{figure}
    \centering
    \includegraphics[width=\linewidth]{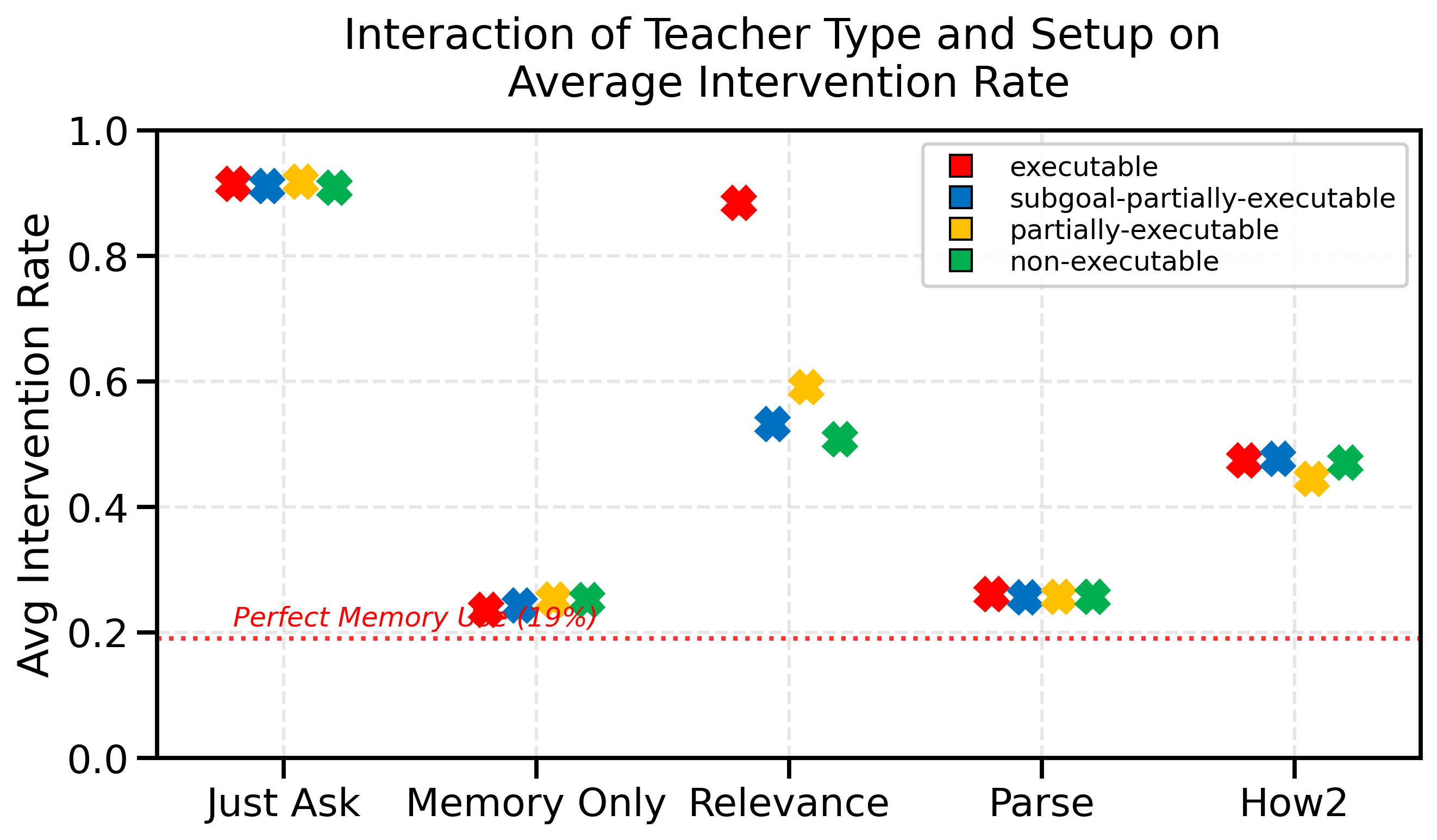}
    \caption{We plot the average intervention rate of the different teachers for all five different memory setups.
        The intervention rate is approximately 0.9 for all teachers in the Just Ask setup.
        The Parse setup achieves the lowest intervention rate, at approximately 0.26, approaching the perfect memory use baseline of 0.19.
        In the Relevance setup, the \executable teacher requires significantly more interventions (0.89) compared to the abstracted teachers (0.51 to 0.59).
        Finally, in the $How^2$ setup, all teachers have a similar intervention rate of around 0.5, which is higher than the Parse setup but significantly lower than the Just Ask setup, indicating the trade-off between asking for help and reusing previous knowledge.
    }
    \label{fig:h1_intervention}
\end{figure}

In Figures~\ref{fig:h1_success} and \ref{fig:h1_intervention}, we plot the success rate and intervention rate of the different teachers for all five different memory setups.
This visualisation of the results supports part 2 of \textbf{H1}: \executable teachers are less reusable than others.

\begin{table}[h!]
    \centering
    \scalebox{0.7}{
        \begin{tabular}{lrrrr}
            \toprule
                                 & sum\_sq & df       & F      & PR(>F) \\
            \midrule
            C(teacher)           & 10.55   & 1.00     & 43.36  & 0.00   \\
            C(memory)            & 45.32   & 1.00     & 186.23 & 0.00   \\
            C(teacher):C(memory) & 19.17   & 1.00     & 78.77  & 0.00   \\
            Residual             & 2744.12 & 11276.00 & NaN    & NaN    \\
            \bottomrule
        \end{tabular}
    }
    \caption{Two-way ANOVA results for success rate, examining the interaction between teacher and memory.}
    \label{tab:anova}
\end{table}

Our second hypothesis (\textbf{H2}) was: abstracting answers into subgoals significantly enhances reusability.
To test \textbf{H2}, we compare the success rates of \subgoalpartiallyexecutable and \partiallyexecutable teachers in all the memory setup (memory-only, parse, relevance, and $How^2$)
The result of the t-test is significant, with a p-value of 1.24-05, indicating that there is a significant difference between the two models in terms of success rate.
This confirms our second hypothesis (\textbf{H2}) and is somewhat surprising as the \subgoalpartiallyexecutable is just a structured version of the \partiallyexecutable teacher.

\subsection{Errors}
\label{app:errors}

In Table~\ref{tab:errors}, we divide errors into three categories: 1) \textit{impossible} errors, where the agent emits impossible but the plan is not impossible; 2) \textit{max steps} errors, where the agent runs out of steps before completing the plan; and 3) \textit{eager crafting} errors, where the agent crafts an item that is not part of the target plan and therefore fails to complete the task.
Overall, we find that agents with teacher access (Just Ask, relevance, $How^2$) make fewer impossible errors but more max steps errors compared to memory-only agents.
This is because they correctly identify impossible tasks more often, but can fail to follow complex plans within the step limit.
For qualitative examples of these errors, see Appendix~\ref{app:qualitative-examples}.

\subsection{Token Usage and Efficiency}
\label{app:token-usage}
In Table~\ref{tab:additional-metrics}, we show the average number of tokens used by the agent in each setup, along with the action efficiency.
We also show the success rate for each setup with the standard deviation in Table~\ref{tab:success-rate}.
As expected, we find that memory-based agents are more token-efficient, especially in the high-repetition setting, as they reuse cached answers instead of querying the teacher.

\begin{table}
    \centering
    \scalebox{0.65}{
        \begin{tabular}{l l cc cc}
            \toprule
                                                       &                              & \multicolumn{2}{c}{\makecell{Token Usage (k)                       \\ ($\downarrow$)}} & \multicolumn{2}{c}{\makecell{Action \\ Efficiency ($\downarrow$)}} \\
                                                       &                              & low                                          & high  & low  & high \\
            \midrule
                                                       & base                         & 20.0k                                        & 19.3k & 1.14 & 1.38 \\
            \midrule
            \multirow{5}{*}{\rotatebox{90}{Just Ask}}  & executable                   & 38.6k                                        & 37.2k & 1.11 & 0.71 \\
                                                       & partially-executable         & 38.3k                                        & 38.4k & 0.63 & 0.32 \\
                                                       & subgoal-partially-executable & 37.1k                                        & 37.6k & 0.63 & 0.55 \\
                                                       & non-executable               & 46.0k                                        & 43.2k & 0.92 & 0.74 \\
            \midrule
            \multirow{5}{*}{\rotatebox{90}{mem}}       & executable                   & 32.2k                                        & 22.6k & 0.80 & 1.15 \\
                                                       & partially-executable         & 31.4k                                        & 23.2k & 0.89 & 0.98 \\
                                                       & subgoal-partially-executable & 30.8k                                        & 22.7k & 0.86 & 0.91 \\
                                                       & non-executable               & 38.7k                                        & 28.1k & 1.49 & 1.13 \\
            \midrule
            \multirow{5}{*}{\rotatebox{90}{parse}}     & executable                   & 33.0k                                        & 24.0k & 0.78 & 0.85 \\
                                                       & partially-executable         & 33.2k                                        & 24.2k & 0.89 & 0.76 \\
                                                       & subgoal-partially-executable & 31.2k                                        & 24.1k & 0.81 & 0.92 \\
                                                       & non-executable               & 33.8k                                        & 24.8k & 1.00 & 0.91 \\
            \midrule
            \multirow{5}{*}{\rotatebox{90}{relevance}} & executable                   & 39.7k                                        & 43.3k & 0.98 & 0.86 \\
                                                       & partially-executable         & 40.0k                                        & 42.8k & 0.69 & 0.61 \\
                                                       & subgoal-partially-executable & 36.8k                                        & 38.0k & 0.69 & 0.66 \\
                                                       & non-executable               & 44.8k                                        & 41.9k & 0.76 & 1.00 \\
            \midrule
            \multirow{5}{*}{\rotatebox{90}{$How^2$}}   & executable                   & 41.4k                                        & 38.4k & 0.67 & 0.60 \\
                                                       & partially-executable         & 41.9k                                        & 37.4k & 0.74 & 0.71 \\
                                                       & subgoal-partially-executable & 39.7k                                        & 37.6k & 0.79 & 0.78 \\
                                                       & non-executable               & 40.2k                                        & 38.9k & 0.83 & 0.87 \\
            \bottomrule
        \end{tabular}}
    \caption{Additional performance metrics of different models. We show the token usage (aggregating the teacher token usage with all the activated agent roles). The action efficiency metric is calculated as in Plancraft \citep{dagan2024plancraftevaluationdatasetplanning}.}
    \label{tab:additional-metrics}
\end{table}

\begin{table*}
    \centering
    \scalebox{0.7}{
        \begin{tabular}{l l cc cc cc cc}
            \toprule
                                                       &                              & \multicolumn{2}{c}{\makecell{Success                                                  \\ Rate ($\uparrow$)}} & \multicolumn{2}{c}{\makecell{Impossible\\Error ($\downarrow$)}} & \multicolumn{2}{c}{\makecell{Max Steps\\Error ($\downarrow$)}} & \multicolumn{2}{c}{\makecell{Eager Crafting\\Error ($\downarrow$)}} \\
                                                       &                              & low                                  & high & low  & high & low  & high & low  & high \\
            \midrule
                                                       & base                         & 0.20                                 & 0.21 & 0.56 & 0.54 & 0.01 & 0.02 & 0.09 & 0.10 \\
            \midrule
            \multirow{5}{*}{\rotatebox{90}{Just Ask}}  & executable                   & 0.59                                 & 0.58 & 0.04 & 0.04 & 0.21 & 0.21 & 0.10 & 0.10 \\
                                                       & partially-executable         & 0.54                                 & 0.53 & 0.05 & 0.05 & 0.23 & 0.24 & 0.10 & 0.11 \\
                                                       & subgoal-partially-executable & 0.57                                 & 0.56 & 0.04 & 0.04 & 0.22 & 0.23 & 0.09 & 0.09 \\
                                                       & non-executable               & 0.50                                 & 0.51 & 0.04 & 0.05 & 0.30 & 0.28 & 0.08 & 0.07 \\
                                                       & avg                          & 0.55                                 & 0.54 & 0.04 & 0.05 & 0.24 & 0.24 & 0.09 & 0.09 \\
            \midrule
            \multirow{5}{*}{\rotatebox{90}{mem}}       & executable                   & 0.43                                 & 0.32 & 0.21 & 0.42 & 0.17 & 0.07 & 0.10 & 0.07 \\
                                                       & partially-executable         & 0.48                                 & 0.41 & 0.17 & 0.34 & 0.16 & 0.08 & 0.10 & 0.07 \\
                                                       & subgoal-partially-executable & 0.52                                 & 0.46 & 0.16 & 0.31 & 0.16 & 0.06 & 0.09 & 0.08 \\
                                                       & non-executable               & 0.44                                 & 0.41 & 0.16 & 0.32 & 0.22 & 0.11 & 0.08 & 0.06 \\
                                                       & avg                          & 0.47                                 & 0.40 & 0.18 & 0.35 & 0.18 & 0.08 & 0.09 & 0.07 \\
            \midrule
            \multirow{5}{*}{\rotatebox{90}{parse}}     & executable                   & 0.48                                 & 0.44 & 0.18 & 0.33 & 0.18 & 0.09 & 0.08 & 0.06 \\
                                                       & partially-executable         & 0.48                                 & 0.43 & 0.17 & 0.33 & 0.18 & 0.08 & 0.09 & 0.06 \\
                                                       & subgoal-partially-executable & 0.51                                 & 0.44 & 0.16 & 0.33 & 0.16 & 0.08 & 0.08 & 0.06 \\
                                                       & non-executable               & 0.49                                 & 0.46 & 0.18 & 0.31 & 0.18 & 0.08 & 0.07 & 0.05 \\
                                                       & avg                          & 0.49                                 & 0.44 & 0.17 & 0.32 & 0.18 & 0.08 & 0.08 & 0.06 \\
            \midrule
            \multirow{5}{*}{\rotatebox{90}{relevance}} & executable                   & 0.58                                 & 0.58 & 0.05 & 0.05 & 0.20 & 0.20 & 0.10 & 0.10 \\
                                                       & partially-executable         & 0.52                                 & 0.50 & 0.06 & 0.08 & 0.23 & 0.24 & 0.11 & 0.10 \\
                                                       & subgoal-partially-executable & 0.55                                 & 0.51 & 0.07 & 0.12 & 0.20 & 0.20 & 0.09 & 0.08 \\
                                                       & non-executable               & 0.46                                 & 0.47 & 0.08 & 0.13 & 0.29 & 0.24 & 0.08 & 0.07 \\
                                                       & avg                          & 0.53                                 & 0.52 & 0.06 & 0.10 & 0.23 & 0.22 & 0.10 & 0.09 \\
            \midrule
            \multirow{5}{*}{\rotatebox{90}{$How^2$}}   & executable                   & 0.52                                 & 0.50 & 0.09 & 0.17 & 0.23 & 0.18 & 0.08 & 0.06 \\
                                                       & partially-executable         & 0.49                                 & 0.49 & 0.09 & 0.18 & 0.24 & 0.17 & 0.09 & 0.07 \\
                                                       & subgoal-partially-executable & 0.53                                 & 0.50 & 0.08 & 0.16 & 0.23 & 0.17 & 0.08 & 0.08 \\
                                                       & non-executable               & 0.53                                 & 0.53 & 0.09 & 0.15 & 0.22 & 0.20 & 0.08 & 0.05 \\
                                                       & avg                          & 0.52                                 & 0.50 & 0.09 & 0.16 & 0.23 & 0.18 & 0.08 & 0.07 \\
            \bottomrule
        \end{tabular}}
    \caption{Error Rates of different Teacher and Memory configurations.}
    \label{tab:errors}
\end{table*}

\begin{table*}
    \centering
    \scalebox{0.7}{
        \begin{tabular}{l l cccc cccc}
            \toprule
                                                       &                              & \multicolumn{8}{c}{Success Rate ($\uparrow$)}                                                                                      \\
            \cmidrule(lr){3-10}
                                                       &                              & \multicolumn{4}{c}{low}                       & \multicolumn{4}{c}{high}                                                           \\
            \cmidrule(lr){3-6} \cmidrule(lr){7-10}
                                                       &                              & Overall                                       & Easy                     & Medium & Hard & Overall          & Easy & Medium & Hard \\
            \midrule
                                                       & base                         & 0.20 ($\pm$0.01)                              & 0.35                     & 0.23   & 0.00 & 0.21 ($\pm$0.01) & 0.32 & 0.34   & 0.00 \\

            \midrule
            \multirow{5}{*}{\rotatebox{90}{Just Ask}}  & executable                   & 0.59 ($\pm$0.01)                              & 0.86                     & 0.70   & 0.20 & 0.58 ($\pm$0.01) & 0.85 & 0.75   & 0.16 \\
                                                       & partially-executable         & 0.54 ($\pm$0.01)                              & 0.87                     & 0.59   & 0.13 & 0.53 ($\pm$0.01) & 0.82 & 0.69   & 0.09 \\
                                                       & subgoal-partially-executable & 0.57 ($\pm$0.02)                              & 0.87                     & 0.62   & 0.18 & 0.56 ($\pm$0.01) & 0.83 & 0.76   & 0.11 \\
                                                       & non-executable               & 0.50 ($\pm$0.01)                              & 0.84                     & 0.51   & 0.09 & 0.51 ($\pm$0.01) & 0.85 & 0.60   & 0.07 \\
                                                       & avg                          & 0.55                                          & 0.86                     & 0.61   & 0.15 & 0.54             & 0.84 & 0.70   & 0.11 \\
            \midrule
            \multirow{5}{*}{\rotatebox{90}{mem}}       & executable                   & 0.43 ($\pm$0.01)                              & 0.71                     & 0.42   & 0.09 & 0.32 ($\pm$0.01) & 0.50 & 0.46   & 0.03 \\
                                                       & partially-executable         & 0.48 ($\pm$0.00)                              & 0.79                     & 0.51   & 0.09 & 0.41 ($\pm$0.02) & 0.64 & 0.60   & 0.03 \\
                                                       & subgoal-partially-executable & 0.52 ($\pm$0.01)                              & 0.81                     & 0.58   & 0.14 & 0.46 ($\pm$0.01) & 0.71 & 0.66   & 0.05 \\
                                                       & non-executable               & 0.44 ($\pm$0.02)                              & 0.76                     & 0.46   & 0.06 & 0.41 ($\pm$0.02) & 0.70 & 0.53   & 0.01 \\
                                                       & avg                          & 0.47                                          & 0.77                     & 0.49   & 0.10 & 0.40             & 0.64 & 0.56   & 0.03 \\
            \midrule
            \multirow{5}{*}{\rotatebox{90}{parse}}     & executable                   & 0.48 ($\pm$0.01)                              & 0.78                     & 0.54   & 0.07 & 0.44 ($\pm$0.01) & 0.71 & 0.59   & 0.03 \\
                                                       & partially-executable         & 0.48 ($\pm$0.02)                              & 0.78                     & 0.56   & 0.07 & 0.43 ($\pm$0.02) & 0.70 & 0.59   & 0.03 \\
                                                       & subgoal-partially-executable & 0.51 ($\pm$0.01)                              & 0.78                     & 0.63   & 0.13 & 0.44 ($\pm$0.01) & 0.66 & 0.67   & 0.05 \\
                                                       & non-executable               & 0.49 ($\pm$0.00)                              & 0.81                     & 0.52   & 0.10 & 0.46 ($\pm$0.02) & 0.76 & 0.59   & 0.04 \\
                                                       & avg                          & 0.49                                          & 0.79                     & 0.56   & 0.09 & 0.44             & 0.71 & 0.61   & 0.03 \\
            \midrule
            \multirow{5}{*}{\rotatebox{90}{relevance}} & executable                   & 0.58 ($\pm$0.01)                              & 0.84                     & 0.69   & 0.20 & 0.58 ($\pm$0.01) & 0.83 & 0.77   & 0.16 \\
                                                       & partially-executable         & 0.52 ($\pm$0.00)                              & 0.83                     & 0.57   & 0.12 & 0.50 ($\pm$0.01) & 0.79 & 0.66   & 0.06 \\
                                                       & subgoal-partially-executable & 0.55 ($\pm$0.00)                              & 0.86                     & 0.59   & 0.17 & 0.51 ($\pm$0.01) & 0.81 & 0.66   & 0.09 \\
                                                       & non-executable               & 0.46 ($\pm$0.02)                              & 0.79                     & 0.49   & 0.07 & 0.47 ($\pm$0.00) & 0.80 & 0.54   & 0.04 \\
                                                       & avg                          & 0.53                                          & 0.83                     & 0.58   & 0.14 & 0.52             & 0.81 & 0.66   & 0.09 \\
            \midrule
            \multirow{5}{*}{\rotatebox{90}{$How^2$}}   & executable                   & 0.52 ($\pm$0.02)                              & 0.83                     & 0.60   & 0.13 & 0.50 ($\pm$0.03) & 0.80 & 0.64   & 0.07 \\
                                                       & partially-executable         & 0.49 ($\pm$0.02)                              & 0.79                     & 0.58   & 0.09 & 0.49 ($\pm$0.02) & 0.78 & 0.63   & 0.08 \\
                                                       & subgoal-partially-executable & 0.53 ($\pm$0.02)                              & 0.81                     & 0.64   & 0.15 & 0.50 ($\pm$0.01) & 0.72 & 0.69   & 0.11 \\
                                                       & non-executable               & 0.53 ($\pm$0.02)                              & 0.87                     & 0.55   & 0.11 & 0.53 ($\pm$0.00) & 0.86 & 0.64   & 0.06 \\
                                                       & avg                          & 0.52                                          & 0.82                     & 0.59   & 0.12 & 0.50             & 0.79 & 0.65   & 0.08 \\
            \bottomrule
        \end{tabular}}
    \caption{Success Rate analysis of different models by complexity.}
    \label{tab:success-rate}
\end{table*}

\section{Cache Misses}
\label{app:cache-misses}
\begin{figure}
    \centering
    \begin{subfigure}[b]{\linewidth}
        \includegraphics[width=\textwidth]{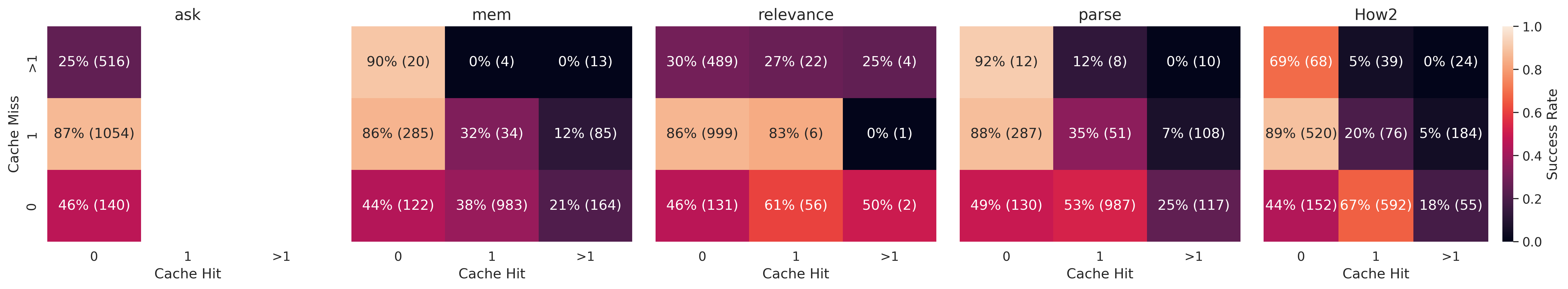}
        \caption{Executable Teacher}
        \label{fig:heatmap-executable1}
    \end{subfigure}
    \hfill %
    \begin{subfigure}[b]{\linewidth}
        \includegraphics[width=\textwidth]{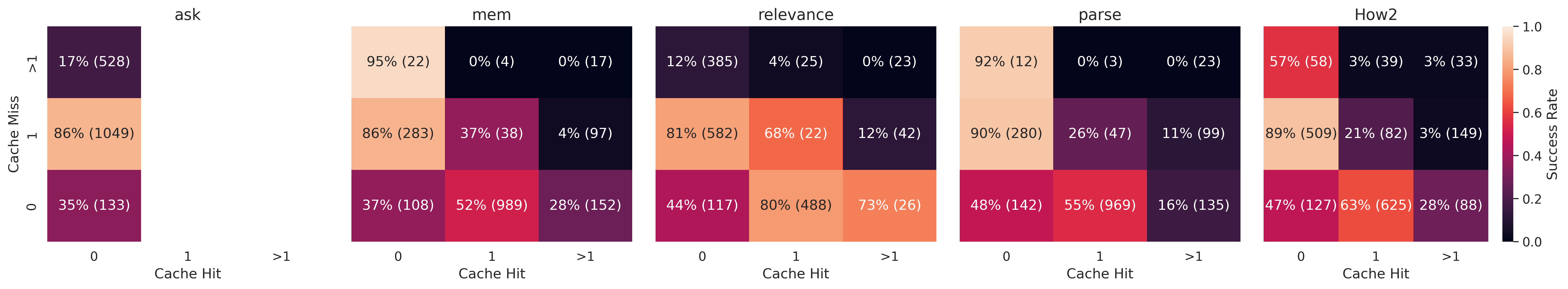}
        \caption{Partially-Executable Teacher}
        \label{fig:heatmap-partially-executable}
    \end{subfigure}
    \vspace{1em} %
    \begin{subfigure}[b]{\linewidth}
        \includegraphics[width=\textwidth]{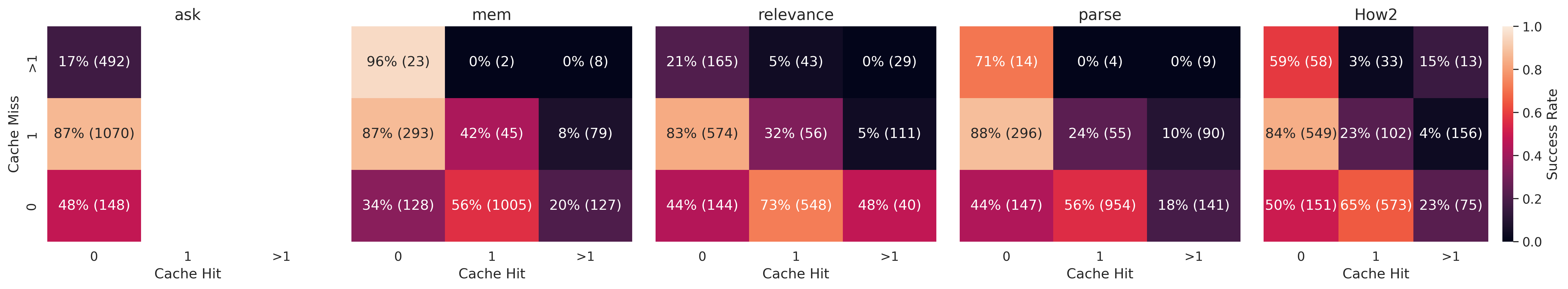}
        \caption{Subgoal-Partially-Executable Teacher}
        \label{fig:heatmap-subgoal}
    \end{subfigure}
    \hfill %
    \begin{subfigure}[b]{\linewidth}
        \includegraphics[width=\textwidth]{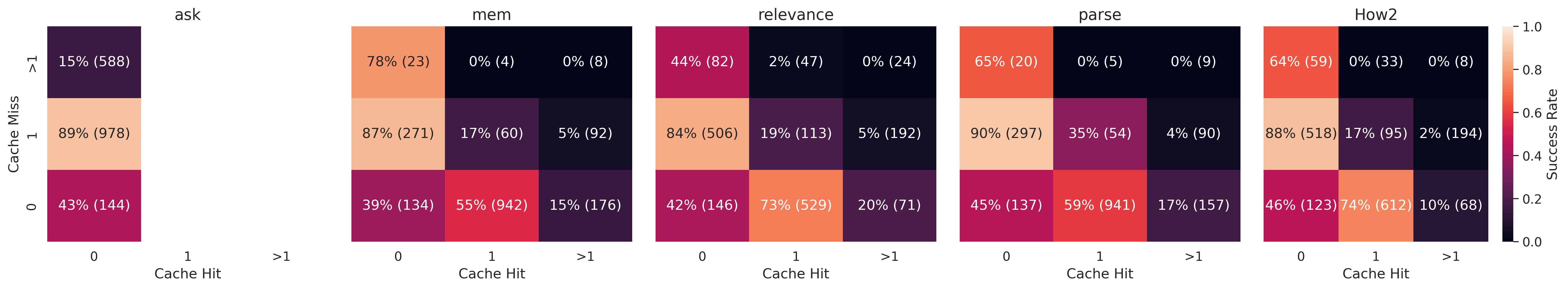}
        \caption{Non-Executable Teacher}
        \label{fig:heatmap-non-executable}
    \end{subfigure}
    \caption{Heatmaps illustrating the performance (Success Rate and Cache Miss Rate) of the $How^2$ framework across different question-asking strategies for each of the four teacher types. The \subgoalpartiallyexecutable{} teacher (c) consistently achieves a strong balance between high success rates and lower cache misses, particularly in the full $How^2$ configuration.}
    \label{fig:all-results-heatmaps}
\end{figure}

In Figure~\ref{fig:all-results-heatmaps}, we show the overall success per cache miss and cache hits for each of the teacher models and setup.
We see the effect of asking questions, as in the Just Ask setup, where the cache hit is one (only one question is asked), most teachers have a high success rate (close to 90\%).
However, if the agent asks more than one question, then this average success rate falls dramatically (from 15\% to 25\%).
This is likely because asking more than one question is caused by agent uncertainty and correlated with task complexity.
We also see that in around 10\% of the cases, the agent does not ask any questions.

Once we move to a memory setup, we can compare the success rate of cache misses with cache hits.
Overall, using memory-only, allows more than two thirds of the answers to be cached, however, the cache hits does not necessarily lead to a higher success rate (38\% to 56\%).
We still find that if the agent asks more than one question (whether or not it uses memory), its success rate drops significantly.
If we use the relevance check setup, we can observe that the number of cache hits decreases (since less memories are deemed relevant), but the overall success rate increases for cache hits.
This is at the cost of asking significantly more questions.
If we parse the teacher's answer, we also improve the accuracy of cache hits and bring all teacher models closer in performance
Finally, the $How^2$ setup, which combines relevance and parsing, achieves high success rate for cache hits while keeping the number of cache misses low.

\begin{table*}
    \centering
    \scalebox{0.65}{
        \begin{tabular}{l l cc cc}
            \toprule
                                                       &                              & \makecell{Overall                                                   \\ SR ($\uparrow$)} & \makecell{Impossible\\F1 ($\uparrow$)} & \makecell{Avg Cache\\Miss ($\downarrow$)} & \makecell{Avg Intervention\\Rate ($\downarrow$)} \\
            \midrule
                                                       & base                         & \underline{0.21}          & \underline{0.40}          & 0.00 & 0.00 \\
            \midrule
            \multirow{5}{*}{\rotatebox{90}{Just Ask}}  & executable                   & 0.43                      & 0.61                      & 0.54 & 0.41 \\
                                                       & partially-executable         & 0.43                      & 0.62                      & 0.52 & 0.40 \\
                                                       & subgoal-partially-executable & 0.48                      & \textbf{\underline{0.62}} & 0.51 & 0.42 \\
                                                       & non-executable               & \underline{0.48}          & 0.62                      & 0.50 & 0.42 \\
                                                       & avg                          & 0.46                      & 0.62                      & 0.52 & 0.41 \\
            \midrule
            \multirow{5}{*}{\rotatebox{90}{mem}}       & executable                   & 0.31                      & 0.50                      & 0.17 & 0.17 \\
                                                       & partially-executable         & 0.37                      & 0.53                      & 0.17 & 0.16 \\
                                                       & subgoal-partially-executable & 0.41                      & \underline{0.55}          & 0.17 & 0.17 \\
                                                       & non-executable               & \underline{0.41}          & 0.55                      & 0.17 & 0.17 \\
                                                       & avg                          & 0.38                      & 0.53                      & 0.17 & 0.17 \\
            \midrule
            \multirow{5}{*}{\rotatebox{90}{parse}}     & executable                   & 0.39                      & 0.54                      & 0.17 & 0.16 \\
                                                       & partially-executable         & 0.39                      & 0.54                      & 0.17 & 0.16 \\
                                                       & subgoal-partially-executable & 0.41                      & \underline{0.55}          & 0.17 & 0.17 \\
                                                       & non-executable               & \underline{0.41}          & 0.54                      & 0.16 & 0.16 \\
                                                       & avg                          & 0.40                      & 0.54                      & 0.17 & 0.16 \\
            \midrule
            \multirow{5}{*}{\rotatebox{90}{relevance}} & executable                   & 0.42                      & \underline{0.61}          & 0.47 & 0.37 \\
                                                       & partially-executable         & 0.43                      & 0.60                      & 0.37 & 0.29 \\
                                                       & subgoal-partially-executable & 0.44                      & 0.59                      & 0.30 & 0.27 \\
                                                       & non-executable               & \underline{0.47}          & 0.60                      & 0.26 & 0.24 \\
                                                       & avg                          & 0.44                      & 0.60                      & 0.35 & 0.29 \\
            \midrule
            \multirow{5}{*}{\rotatebox{90}{$How^2$}}   & executable                   & 0.45                      & 0.60                      & 0.32 & 0.27 \\
                                                       & partially-executable         & 0.44                      & 0.60                      & 0.32 & 0.26 \\
                                                       & subgoal-partially-executable & 0.47                      & 0.60                      & 0.27 & 0.25 \\
                                                       & non-executable               & \textbf{\underline{0.49}} & \underline{0.62}          & 0.29 & 0.25 \\
                                                       & avg                          & 0.46                      & 0.60                      & 0.30 & 0.26 \\
            \bottomrule
        \end{tabular}}
    \caption{Performance of the Qwen3 32B model on the high-repetition split. We report Success Rate (SR), Impossible F1-score, Average Cache Miss Rate, and Average Intervention Rate.}
    \label{tab:qwen-results}
\end{table*}

\section{Reasoning Model}
\label{app:qwen}
We also run our experiments on the high-repetition split using the Qwen 3 32B model \citep{yang2025qwen3technicalreport}, which is specifically fine-tuned for reasoning.
For this setup, we remove the explicit \texttt{think} action, as the model is trained to generate a reasoning trace before each action.
We enable this implicit reasoning for all agent roles and the \nonexecutable teacher, re-using the same prompts as for the Llama 3.3 70B model.

The results are presented in Table~\ref{tab:qwen-results}.
Overall, Qwen 3 is less effective than Llama 3.3, with the average success rate for the full $How^2$ framework dropping from 0.52 to 0.46.
The primary cause for this performance degradation appears to be a lower frequency of invoking the \texttt{read-memory} action.
For instance, in the $How^2$ setup, Qwen 3's average intervention rate is 0.26, compared to 0.53 for Llama 3.3.
This suggests that the reasoning trace may create a bias towards immediate action (e.g., moving or smelting) rather than information-seeking through memory retrieval.

This experiment is nonetheless valuable, as it demonstrates how our framework adapts to a model that relies less on external guidance.
In this context, $How^2$ is particularly effective, achieving an average success rate of 0.46, which is on par with the \textit{Just Ask} setup (0.46), but with at significantly lower intervention rate (0.26 vs. 0.41).
This shows that $How^2$ can successfully leverage a limited number of teacher interactions for effective learning.

A notable difference from the Llama 3.3 experiments is that the \nonexecutable teacher consistently yields the best performance with Qwen 3 when compared to other teachers, achieving a success rate of 0.49 in the $How^2$ setup.
This may be because the reasoning capabilities of the Qwen 3 backbone, used by both the agent and the teacher, are better suited to generating and interpreting abstract, ungrounded instructions.
The teacher model can produce more effective high-level plans, and the agent's reasoning trace allows it to better parse and execute these plans.
As a result, we find contradictory results to the first part of \textbf{H1} that the \executable teacher is the most immediately useful, but find support for the second part of \textbf{H1} that the \executable teacher is the least reusable.

As for \textbf{H2}, we find that the \subgoalpartiallyexecutable teacher is still the most reusable compared to the \partiallyexecutable teacher, with a success rate of 0.47 compared to 0.44 in the $How^2$ setup.
This supports our hypothesis that abstracting answers into subgoals significantly enhances reusability, even with a different LLM.

%% file: prompts.tex
\section{Prompts}
\label{app:prompts}
This section contains the prompts used in the \texorpdfstring{$How^2$}{How2} framework.
As mentioned, we follow the tool-call paradigm, where each action is defined as a JSON tool call, and follow the recommended prompt format for each of the models we test.
The JSON schema for each tool is provided below in Figures~\ref{fig:read-memory-tool}, \ref{fig:think-tool}, \ref{fig:move-tool}, \ref{fig:smelt-tool}, \ref{fig:impossible-tool}.

The main system prompt, which defines the environment rules and action constraints within Plancraft, and is used by the main agent, the Parser, Relevance Check and Ask roles in the $How^2$ framework, is shown in Figure~\ref{fig:main-prompt}.
This system prompt is similar to the one used in \citet{dagan2024plancraftevaluationdatasetplanning}, the main difference being that specific actions are defined as JSON tool calls and the parameter space of slots does not use the square bracket notation \highlightB{[IXX]} and instead refers to slots directly as \highlightB{IXX}.
We change the slot parameter space to be more friendly to JSON tool-call generation, as we found models struggle to generate the square bracket notation in a consistent manner when generating tool calls with the bracket notation (\highlightB{[IXX]}).
The additional description of the \highlightG{Memory System} is only added to the system prompt of the Parser, Relevance Check and Ask roles, as these roles interact with the memory system directly.

Figures~\ref{fig:relevance-prompt}, \ref{fig:ask-prompt} and \ref{fig:parse-prompt} show the prompts for the Relevance Check, Ask and Parse roles respectively.
The Relevance Check role checks if a cached memory entry is applicable to the current game state, the Ask role formulates a procedural `how-to` question when it encounters a knowledge gap, and the Parse role structures the teacher's answer into a generalised format suitable for long-term storage and reuse

The Teacher prompts for the \nonexecutable teacher is shown in Figures~\ref{fig:non-executable-teacher-prompt}.
The placeholder for \highlightR{\{\{context\}\}} and \highlightB{\{\{planner\_str\}\}} are dynamically filled with the latest game state observation, providing the necessary context for the agent or teacher to perform its task.
Since we wish the teacher to be ungrounded in the specific inventory state, to prevent leakage of specific slot placement, we remove specific slot placements from the observation (context) and planner output.

\begin{figure*}[h!]
    \begin{minipage}{\textwidth}
        \begin{tcolorbox}[title={System Prompt}, colbacktitle=bluePP, unbreakable]
            You are crafting in Minecraft. You need to decide on the next action. \\
            \newline
            Crafting Grid: The crafting table is organized into a 3x3 grid. Each slot in the grid has a unique identifier: \\
            - Top row: A1 A2 A3 \\
            - Middle row: B1 B2 B3 \\
            - Bottom row: C1 C2 C3 \\
            \newline
            The output of the crafting process is placed in a designated output slot labeled 0 \\
            You cannot move or smelt items directly into slot 0 \\
            \newline
            The remaining inventory slots (outside of the crafting grid) are used for storing items. These slots are labeled as I1 to I36 \\
            \newline
            Before you take an action, try to think and plan the intermediate goals and next steps to take. \\
            \newline
            Constraints: \\
            - You cannot move or smelt items into slot 0 \\
            - If an item is not in slot 0 then the recipe is incorrect \\
            - You need to move items from slot 0 to a free inventory slot to complete the crafting process \\
            - If you move an item to a slot already containing an item, nothing will happen \\
            - Only call one tool or action at a time \\
            \newline
            \highlightG{Memory System:\\
                - You have access to a memory system where you can store and retrieve recipes and crafting knowledge.\\
                - Your goal is to learn recipes and procedures by asking procedural questions and organizing the answers for future use.\\
                - When storing information to memory, structure it consistently with clear component lists and steps.\\
                - When checking relevance, consider if the memory is applicable to the current crafting situation.}
        \end{tcolorbox}
    \end{minipage}
    \caption{The system prompt, defining the environment rules and action constraints within Plancraft and used by the main agent, the Parser, Relevance Check and Ask roles in the $How^2$ framework.
    This system prompt is similar to the one used in \citet{dagan2024plancraftevaluationdatasetplanning}, the main difference being that specific actions are defined as JSON tool calls and the parameter space of slots does not use the square bracket notation `[IXX]' and instead refers to slots directly as `IXX'.
    We change the slot parameter space to be more friendly to JSON tool-call generation, as we found models struggle to generate the square bracket notation in a consistent manner when generating tool calls.
    The additional description of the \highlightG{Memory System} is only added to the system prompt of the Parser, Relevance Check and Ask roles, as these roles interact with the memory system directly.
    }
    \label{fig:main-prompt}
\end{figure*}

\begin{figure*}[h!]
    \begin{minipage}{\textwidth}
        \begin{tcolorbox}[title={Read Memory Tool}, colbacktitle=bluePP, unbreakable]
            \begin{verbatim}
{
  "type": "function",
  "function": {
    "name": "read_memory",
    "description": "Search a database to retrieve memories/instructions for a given recipe.
                    Call this first or if unsure.",
    "parameters": {
      "type": "object",
      "properties": {
        "recipe": {
          "type": "string",
          "description": "Recipe name to search in memory."
        }
      },
      "required": ["recipe"]
    }
  }
}
            \end{verbatim}
        \end{tcolorbox}
    \end{minipage}
    \caption{The JSON schema for the \texttt{read\_memory} tool. This tool allows the agent to search the memory database for previously stored recipes and instructions.}
    \label{fig:read-memory-tool}
\end{figure*}

\begin{figure*}[h!]
    \begin{minipage}{\textwidth}
        \begin{tcolorbox}[title={Think Tool}, colbacktitle=bluePP, unbreakable]
            \begin{verbatim}
{
  "type": "function",
  "function": {
    "name": "think",
    "description": "Generate thoughts to help you decide on the next action",
    "parameters": {
      "type": "object",
      "properties": {
        "thought": {
          "type": "string",
          "description": "<thought message>"
        }
      },
      "required": ["thought"]
    }
  }
}
            \end{verbatim}
        \end{tcolorbox}
    \end{minipage}
    \caption{The JSON schema for the \texttt{think} tool. This tool enables the agent to generate internal thoughts to guide its decision-making process.}
    \label{fig:think-tool}
\end{figure*}

\begin{figure*}[h!]
    \begin{minipage}{\textwidth}
        \begin{tcolorbox}[title={Move Tool}, colbacktitle=bluePP, unbreakable]
            \begin{verbatim}
{
  "type": "function",
  "function": {
    "name": "move",
    "description": "Transfer a specific quantity of an item from one slot to another",
    "parameters": {
      "type": "object",
      "properties": {
        "slot_from": {
          "type": "string",
          "description": "The slot to move the item from (A1, A2, B1, B2, etc.)"
        },
        "slot_to": {
          "type": "string",
          "description": "The slot to move the item to (A1, A2, B1, B2, etc.)"
        },
        "quantity": {
          "type": "integer",
          "description": "The number of items to move"
        }
      },
      "required": ["slot_from", "slot_to", "quantity"]
    }
  }
}
            \end{verbatim}
        \end{tcolorbox}
    \end{minipage}
    \caption{The JSON schema for the \texttt{move} tool. This tool allows the agent to move items between different slots in the crafting grid or inventory.}
    \label{fig:move-tool}
\end{figure*}

\begin{figure*}[h!]
    \begin{minipage}{\textwidth}
        \begin{tcolorbox}[title={Smelt Tool}, colbacktitle=bluePP, unbreakable]
            \begin{verbatim}
{
    "type": "function",
    "function": {
        "name": "smelt",
        "description": "Smelt an item in a furnace and moves the output to a specific slot",
        "parameters": {
            "type": "object",
            "properties": {
                "slot_from": {
                    "type": "string",
                    "description": "The slot to smelt the item from (A1, A2, B1, B2, etc.)"
                },
                "slot_to": {
                    "type": "string",
                    "description": "The slot to smelt the item to (A1, A2, B1, B2, etc.)"
                },
                "quantity": {
                    "type": "integer",
                    "description": "The number of items to smelt"
                }
            },
            "required": ["slot_from", "slot_to", "quantity"]
        }
    }
}
\end{verbatim}
        \end{tcolorbox}
    \end{minipage}
    \caption{The JSON schema for the \texttt{smelt} tool. This tool enables the agent to smelt items in Plancraft.}
    \label{fig:smelt-tool}
\end{figure*}

\begin{figure*}[h!]
    \begin{minipage}{\textwidth}
        \begin{tcolorbox}[title={Impossible Tool}, colbacktitle=bluePP, unbreakable]
            \begin{verbatim}
{
    "type": "function",
    "function": {
        "name": "impossible",
        "description": "Stop task if it is certain that it is impossible with given inventory",
        "parameters": {
            "type": "object",
            "properties": {
                "reason": {
                    "type": "string",
                    "description": "The reason why the action is impossible"
                }
            },
            "required": ["reason"]
        }
    }
}
\end{verbatim}
        \end{tcolorbox}
    \end{minipage}
    \caption{The JSON schema for the \texttt{impossible} tool. This tool allows the agent to declare when a task cannot be completed, providing a reason for the impossibility.}
    \label{fig:impossible-tool}
\end{figure*}

\begin{figure*}[h!]
    \begin{tcolorbox}[title={Ask Prompt}, breakable, colbacktitle=bluePP, unbreakable]
        \# Agent Context: \\
        \highlightR{\{\{context\}\}}\\
        \\
        Formulate a procedural how-to question about \highlightY{\{\{recipe\_name\}\}}\\
        \\
        Consider the following:\\
        1. Focus on asking **how** to craft the target item mentioned in the Agent Context\\
        2. Use the agent context\\
        3. Ensure the question is concise and focused on \highlightY{\{\{recipe\_name\}\}}\\

        Based on the above, please provide a clear, well-formed question about \highlightY{\{\{recipe\_name\}\}}.\\
        Only return the question. Do not include any additional text or context.\\
    \end{tcolorbox}
    \caption{The prompt for the `ask' role. This guides the agent in formulating a procedural `how-to` question when it encounters a knowledge gap.}
    \label{fig:ask-prompt}
\end{figure*}

\begin{figure*}[h!]
    \begin{tcolorbox}[title={Relevance Check Prompt}, breakable, colbacktitle=bluePP, unbreakable]
        \# Agent Context: \\
        \\
        \highlightR{\{\{context\}\}}\\
        \\
        \# Memory for ``\highlightY{\{\{recipe\_name\}\}}''\\
        \\
        \highlightG{\{\{memory\}\}}\\
        \\
        Check if the memory is relevant to the goal and inventory.\\
        \\
        Answer with ``yes'' if one or more of the listed recipes can be used in this context. Otherwise, answer with ``no''.\\
        Do not include any additional text or context in your response. \\
    \end{tcolorbox}
    \caption{The prompt for the `relevance-check` role. This is used to validate whether a cached memory entry is applicable to the current game state.}
    \label{fig:relevance-prompt}
\end{figure*}

\begin{figure*}[h!]
    \begin{tcolorbox}[title={Parse Prompt}, breakable, colbacktitle=bluePP, unbreakable]
        Format the Teacher Answer for storage in memory. \\
        \\
        Where the RECIPE is the primary item, recipe or concept being described. \\
        The original recipe name (\highlightY{\{\{recipe\_name\}\}}) can be renamed if necessary to be more specific. \\
        Avoid generic terms like "item", "recipe", "crafting", "object", "inventory", etc. in the RECIPE name. \\
        \\
        Structure your memory entry with these sections:\\
        RECIPE: \highlightY{\{\{recipe\_name\}\}}\\
        REQUIREMENTS: [list of prequisites or materials]\\
        PROCEDURE: [step-by-step instructions]\\
        RELATED ITEMS: [list relevant related recipe items in python list format]\\
        \\
        Keep your entry concise and focused on information contained within the Teacher answer.
        \\
        \# Agent Context \\
        \highlightR{\{\{context\}\}} \\
        \\
        \# Agent Question \\
        \highlightY{\{\{question\}\}} \\
        \\
        \# Teacher's Answer \\
        \highlightB{\{\{answer\}\}} \\
        \\
        Format the Teacher's answer into a well-structured memory entry such that it will be useful for future interactions.
    \end{tcolorbox}
    \caption{The prompt for the `parse` role. This structures the teacher's answer into a generalised format suitable for long-term storage and reuse.}
    \label{fig:parse-prompt}
\end{figure*}

\begin{figure*}[ht]
    \begin{tcolorbox}[title={Non-executable Teacher Prompt}, colbacktitle=bluePP, unbreakable]
        You are an expert Minecraft mentor providing high-level guidance on crafting tasks in answer to user questions.\\
        \newline
        Given the Agent Context and Planner Output, answer the user question.\\
        \newline
        \# Background:\\
        Crafting Grid: The crafting table is organized into a 3x3 grid.\\
        The output of the crafting process is placed in a designated output slot\\
        One cannot move or smelt items directly into slot the output slot\\
        The remaining inventory slots (outside of the crafting grid) are used for storing items.\\
        \newline
        \# Usage Notes:\\
        - Smelting is done with the `smelt` command (no furnace needed)\\
        - The crafting grid is already open (do not instruct to open it)\\
        \newline
        \# Answer Guidelines:\\
        - Use simple language and split the problem into the steps outlined in the Planner Output\\
        - Organize information in a logical sequence\\
        - The Planner Output is **always** correct\\
        - Use shapes (e.g. 2x2) or patterns to describe the arrangement of items in the crafting grid\\
        \newline
        \# Output Format: \\
        Brief abstract explanation of each required step and its components in a paragraph format.  \\
        \newline
        \# Example
        question: How can I craft a glass\_bottle given my inventory? \\
        answer: To craft a glass\_bottle, first smelt sand to obtain three glass items, then arrange the glass in a small V shape in the crafting grid. \\
        \newline
        \# Agent Context\\
        \highlightR{\{\{context\}\}} \\
        \newline
        \# Planner Output \\
        \highlightB{\{\{planner\_str\}\}}
    \end{tcolorbox}
    \caption{The prompt for the non-executable teacher. This prompt instructs the teacher model to provide high-level, conceptual guidance based on the agent's context and a planner's output, without giving away a directly executable sequence of actions.}
    \label{fig:non-executable-teacher-prompt}
\end{figure*}

%% file: qualitative.tex
\section{Qualitative Examples}
\label{app:qualitative-examples}

\begin{figure*}[h!]
    \begin{tcolorbox}[title={Success: VAL0531 (Target: crimson\_planks)}, colbacktitle=greenPP, unbreakable]
        \small
        \begin{minipage}{\textwidth}
            \begin{minipage}{0.65\textwidth}
                \textbf{Dialogue:}\\
                \textcolor{black}{\textbf{User:}} Craft an item of type: crimson\_planks\\
                inventory:\\
                - mooshroom\_spawn\_egg I7 quantity 14\\
                - netherite\_ingot I12 quantity 5...\\
                \vspace{0.5em}
                \textcolor{yellowPP}{\textbf{read\_memory(\{"recipe": "crimson\_planks"\})}}\\
                \vspace{0.5em}
                \textcolor{redPP}{\textbf{Cache Miss}}\\
                \vspace{0.5em}
                \textcolor{greenPP}{\textbf{Question}} How do I craft crimson\_planks?\\
                \vspace{0.5em}
                \textcolor{greenPP}{\textbf{Answer:}} To craft a crimson\_planks, follow these steps:\\
                1. Craft crimson\_planks\\
                1.1. move crimson\_hyphae to A1\\
                1.2. move crimson\_planks to a free inventory slot\\
                \\
                \vspace{0.5em}
                \textcolor{yellowPP}{\textbf{move(\{"slot\_from": "I15", "slot\_to": "A1", "quantity": 1\})}}\\
                \vspace{0.5em}
                \textcolor{black}{\textbf{User:}} Craft an item of type: crimson\_planks\\
                inventory:\\
                - crimson\_planks 0 quantity 4\\
                - crimson\_hyphae A1 quantity 1\\
                - moosh...\\
                \vspace{0.5em}
                \textcolor{yellowPP}{\textbf{move(\{"slot\_from": "0", "slot\_to": "I1", "quantity": 4\})}}\\
                \vspace{0.5em}
                \textcolor{greenPP}{\checkmark~\textbf{Crafting succeeded!}}
            \end{minipage}
            \hfill
            \begin{minipage}{0.32\textwidth}
                \textbf{Environment Frames:}\\
                \textbf{Frame 1:}\\
                \includegraphics[width=\textwidth]{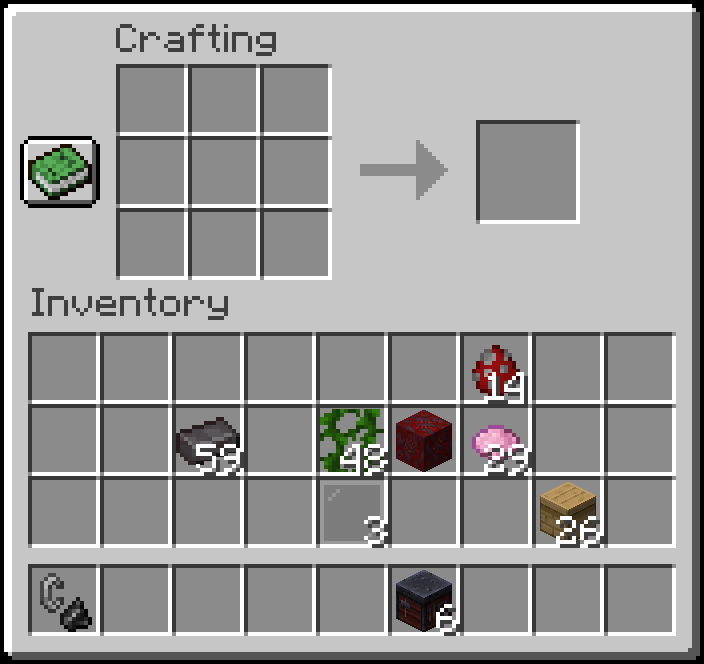}\\
                \vspace{0.5em}
                \textbf{Frame 2:}\\
                \includegraphics[width=\textwidth]{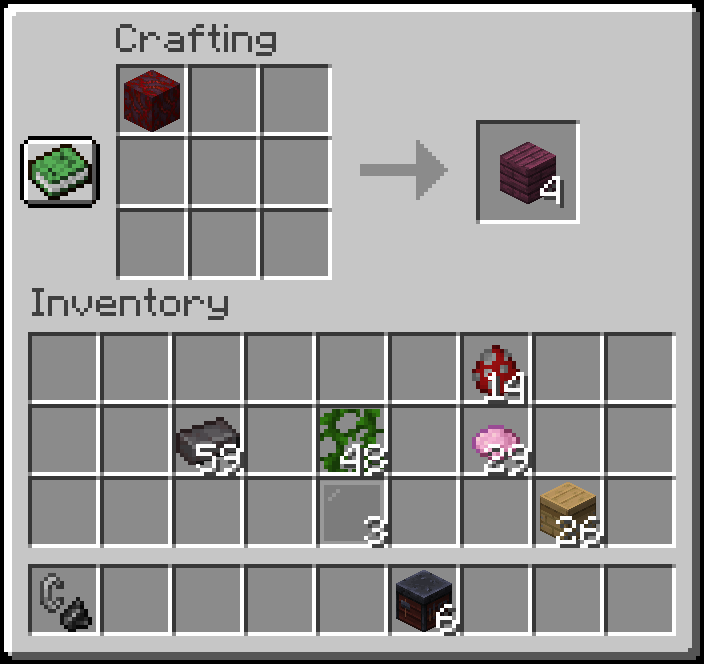}\\
                \vspace{0.5em}
            \end{minipage}
        \end{minipage}

    \end{tcolorbox}
    \caption{Success example from Just Ask and the \subgoalpartiallyexecutable teacher. Memory is read, and since we are in Just Ask setup, the agent asks a how-to question to the teacher. The result is a plan of actions (two actions required) to craft the goal (crimson planks). The Actor is able to successfully follow the Teacher's plan grounding the item name to specific positions.}
    \label{fig:qualitative_ask_subgoal_partially_executable_success_VAL0531}
\end{figure*}

\begin{figure*}[h!]
    \begin{tcolorbox}[title={Failure: VAL0540 (Target: crafting\_table)}, colbacktitle=redPP, unbreakable]
        \small
        \begin{minipage}{\textwidth}
            \begin{minipage}{0.65\textwidth}
                \textbf{Dialogue:}\\
                \textcolor{black}{\textbf{User:}} Craft an item of type: crafting\_table\\
                inventory:\\
                - elder\_guardian\_spawn\_egg I1 quantity 5\\
                - blue\_bed I2 quantity 1\\
                - ...\\
                \vspace{0.5em}
                \textcolor{yellowPP}{\textbf{think(\{"thought": "To craft a crafting table, I need 4 oak logs. I only have 1 oak log in my inventory. I need to find a way to get more oak logs."\})}}\\
                \vspace{0.5em}
                \textcolor{yellowPP}{\textbf{impossible(\{"reason": "not enough oak logs in inventory to craft a crafting table"\})}}\\
                \vspace{0.5em}
                \textcolor{redPP}{\texttimes~\textbf{Crafting failed!}}
            \end{minipage}
            \hfill
            \begin{minipage}{0.32\textwidth}
                \textbf{Environment Frames:}\\
                \textbf{Frame 1:}\\
                \includegraphics[width=\textwidth]{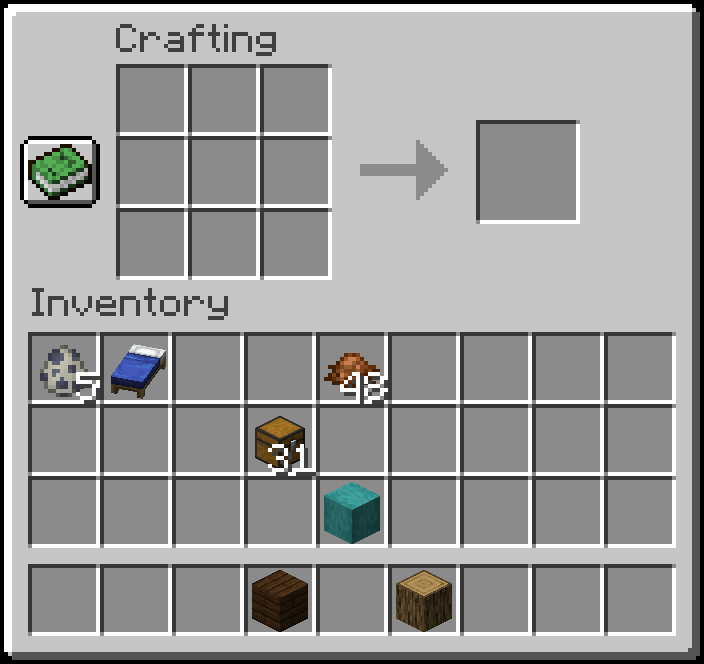}\\
                \vspace{0.5em}
            \end{minipage}
        \end{minipage}

    \end{tcolorbox}
    \caption{Failure example from Just Ask and the \subgoalpartiallyexecutable teacher. The agent emits an impossible action even though the task is solvable. The agent should have recognised that the oak logs can be crafted into planks, which are the main ingredient for the crafting table. In this case, the agent fails to call the memory module and therefore no teacher is consulted. }
    \label{fig:qualitative_ask_subgoal_partially_executable_failure_VAL0540}
\end{figure*}

\begin{figure*}[h!]
    \begin{tcolorbox}[title={Success: VAL0336 (Target: lime\_wool)}, colbacktitle=greenPP, unbreakable]
        \small
        \begin{minipage}{\textwidth}
            \begin{minipage}{0.65\textwidth}
                \textbf{Dialogue:}\\
                \textcolor{black}{\textbf{User:}} Craft an item of type: lime\_wool\\
                inventory:\\
                - jungle\_stairs I2 quantity 45\\
                - dark\_oak\_fence I3 quantity 37\\
                - pink\_car...\\
                \vspace{0.5em}
                \textcolor{yellowPP}{\textbf{read\_memory(\{"recipe": "lime\_wool"\})}}\\
                \vspace{0.5em}
                \textcolor{redPP}{\textbf{Cache Miss}}\\
                \vspace{0.5em}
                \textcolor{greenPP}{\textbf{Question}} How do I craft lime\_wool?\\
                \vspace{0.5em}
                \textcolor{greenPP}{\textbf{Answer:}} To craft a lime\_wool, follow these steps:\\
                1. move: from I7 to A1 with quantity 1\\
                2. move: from I15 to A2 with quantity 1\\
                3. move: from 0 to I1 with quantity 1\\
                \\
                \vspace{0.5em}
                \textcolor{yellowPP}{\textbf{move(\{"slot\_from": "I7", "slot\_to": "A1", "quantity": 1\})}}\\
                \vspace{0.5em}
                \textcolor{black}{\textbf{User:}} Craft an item of type: lime\_wool\\
                inventory:\\
                - lime\_dye A1 quantity 1\\
                - jungle\_stairs I2 quantity 45\\
                - dark\_oak\_fence ...\\
                \vspace{0.5em}
                \textcolor{yellowPP}{\textbf{move(\{"slot\_from": "I15", "slot\_to": "A2", "quantity": 1\})}}\\
                \vspace{0.5em}
                \textcolor{black}{\textbf{User:}} Craft an item of type: lime\_wool\\
                inventory:\\
                - lime\_wool 0 quantity 1\\
                - lime\_dye A1 quantity 1\\
                - white\_wool A2 quantit...\\
                \vspace{0.5em}
                \textcolor{yellowPP}{\textbf{move(\{"slot\_from": "0", "slot\_to": "I1", "quantity": 1\})}}\\
                \vspace{0.5em}
                \textcolor{greenPP}{\checkmark~\textbf{Crafting succeeded!}}
            \end{minipage}
            \hfill
            \begin{minipage}{0.32\textwidth}
                \textbf{Environment Frames:}\\
                \textbf{Frame 1:}\\
                \includegraphics[width=\textwidth]{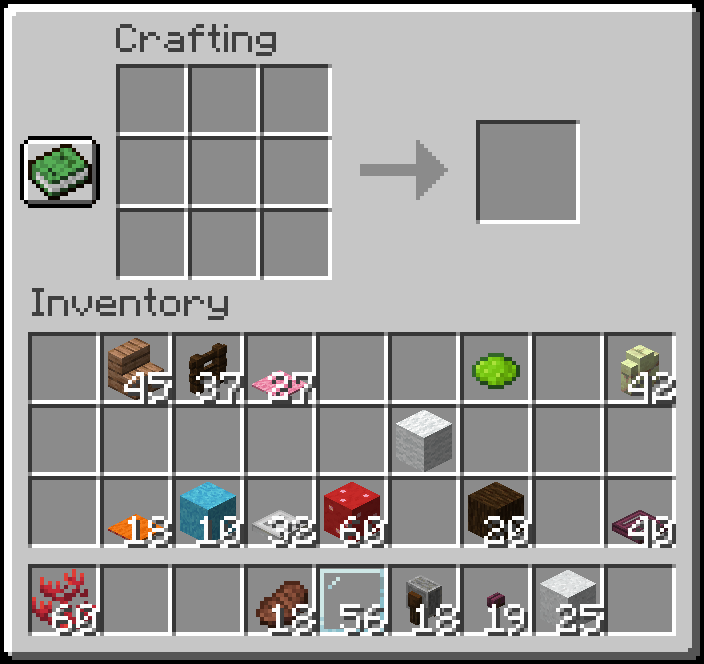}\\
                \vspace{0.5em}
                \textbf{Frame 2:}\\
                \includegraphics[width=\textwidth]{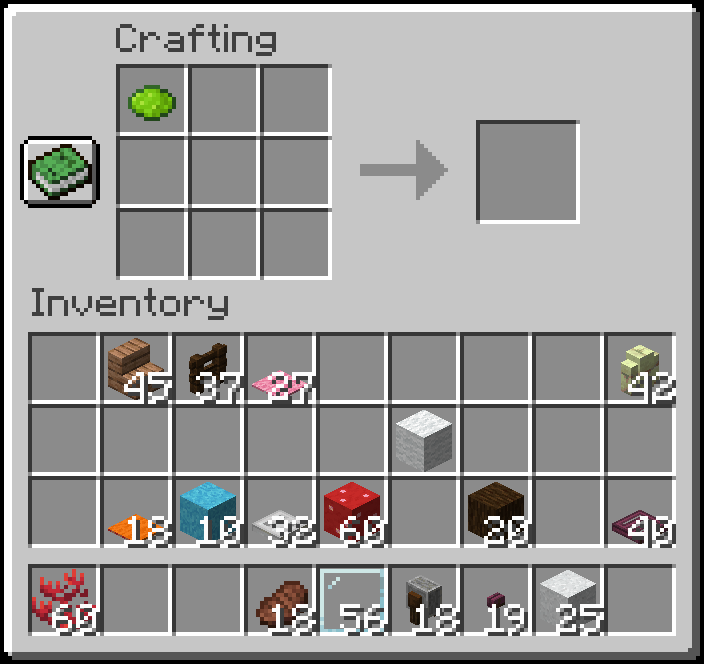}\\
                \vspace{0.5em}
                \textbf{Frame 3:}\\
                \includegraphics[width=\textwidth]{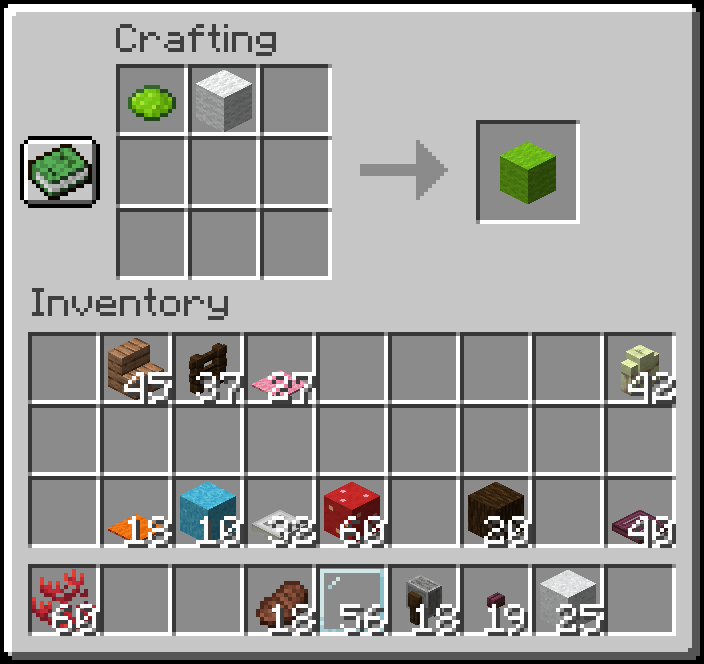}\\
                \vspace{0.5em}
            \end{minipage}
        \end{minipage}

    \end{tcolorbox}
    \caption{Success example a from Memory-Only \executable teacher. Compared to the partially executable plans, the \executable teacher provides fully grounded answers.}
    \label{fig:qualitative_mem_executable_success_VAL0336}
\end{figure*}

\begin{figure*}[h!]
    \begin{tcolorbox}[title={Failure: VAL0356 (Target: oak\_boat)}, colbacktitle=redPP, unbreakable]
        \small
        \begin{minipage}{\textwidth}
            \begin{minipage}{0.65\textwidth}
                \textbf{Dialogue:}\\
                \textcolor{black}{\textbf{User:}} Craft an item of type: oak\_boat\\
                inventory:\\
                - turtle\_spawn\_egg I3 quantity 19\\
                - orange\_bed I4 quantity 1\\
                - smooth\_ston...\\
                \vspace{0.5em}
                \textcolor{yellowPP}{\textbf{read\_memory(\{"recipe": "oak\_boat"\})}}\\
                \vspace{0.5em}
                \textcolor{greenPP}{\textbf{Memory (Cache Hit):}}To craft a oak\_boat, follow these steps:\\
                1. move: from I10 to A1 with quantity 1\\
                2. move: from 0 to I1 with quantity 4\\
                3. move: from I10 to A1 with quantity 1\\
                4. move: from 0 to I1 with quantity 4\\
                5. ...\\
                \vspace{0.5em}
                \textcolor{yellowPP}{\textbf{think(\{"thought": "I do not have oak\_planks in slot I10, I have them in slot I20. I will move oak\_planks from slot I20 to slot A1 with quantity 1."\})}}\\
                \vspace{0.5em}
                \textcolor{black}{\textbf{User:}} Craft an item of type: oak\_boat\\
                inventory:\\
                - turtle\_spawn\_egg I3 quantity 19\\
                - orange\_bed I4 quantity 1\\
                - smooth\_ston...\\
                \vspace{0.5em}
                \textcolor{yellowPP}{\textbf{move(\{"slot\_from": "I20", "slot\_to": "A1", "quantity": 1\})}}\\
                \vspace{0.5em}
                \textcolor{black}{\textbf{User:}} Craft an item of type: oak\_boat\\
                inventory:\\
                - oak\_button 0 quantity 1\\
                - oak\_planks A1 quantity 1\\
                - turtle\_spawn\_egg I3...\\
                \vspace{0.5em}
                \textcolor{yellowPP}{\textbf{think(\{"thought": "I need to move the crafted oak\_button from slot 0 to a free inventory slot to complete the crafting process."\})}}\\
                \vspace{0.5em}
                \textcolor{yellowPP}{\textbf{move(\{"slot\_from": "0", "slot\_to": "I1", "quantity": 1\})}}\\
                \vspace{0.5em}
                \textcolor{redPP}{\texttimes~\textbf{Crafting failed!}}
            \end{minipage}
            \hfill
            \begin{minipage}{0.32\textwidth}
                \textbf{Environment Frames:}\\
                \textbf{Frame 1:}\\
                \includegraphics[width=\textwidth]{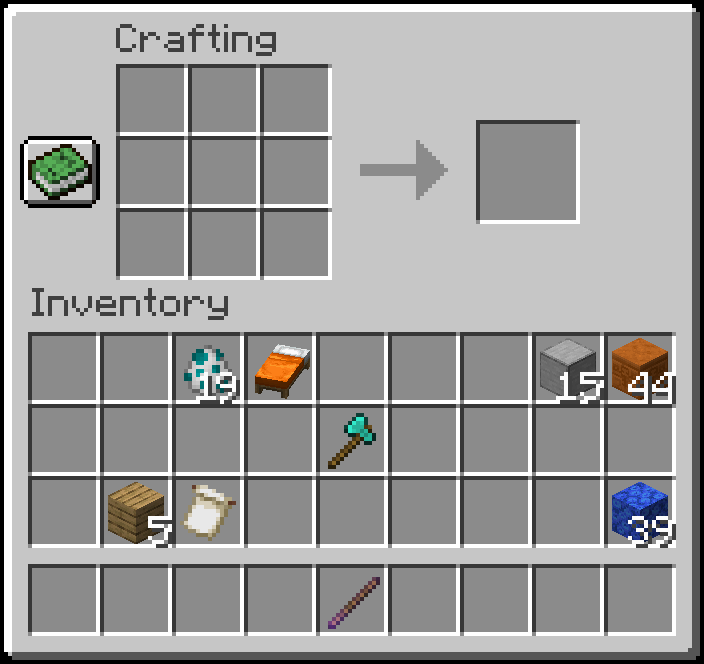}\\
                \vspace{0.5em}
                \textbf{Frame 2:}\\
                \includegraphics[width=\textwidth]{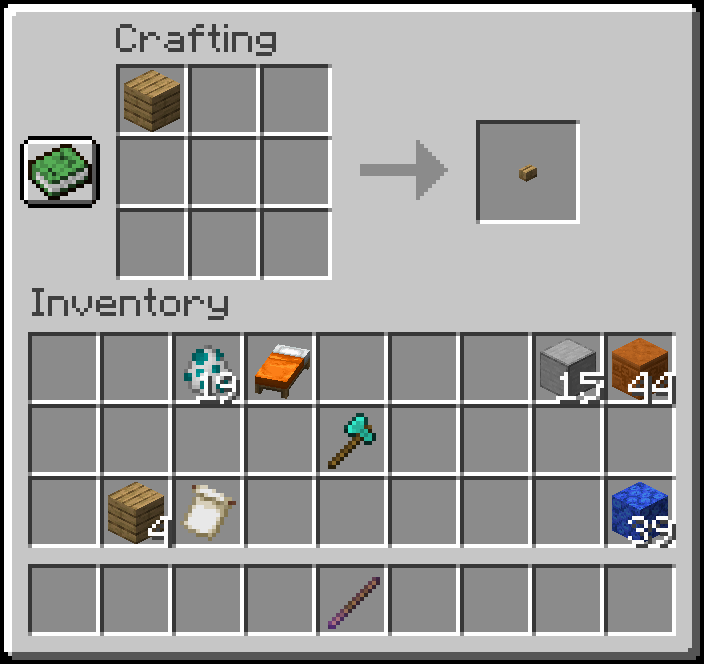}\\
                \vspace{0.5em}
                \textbf{Frame 3:}\\
                \includegraphics[width=\textwidth]{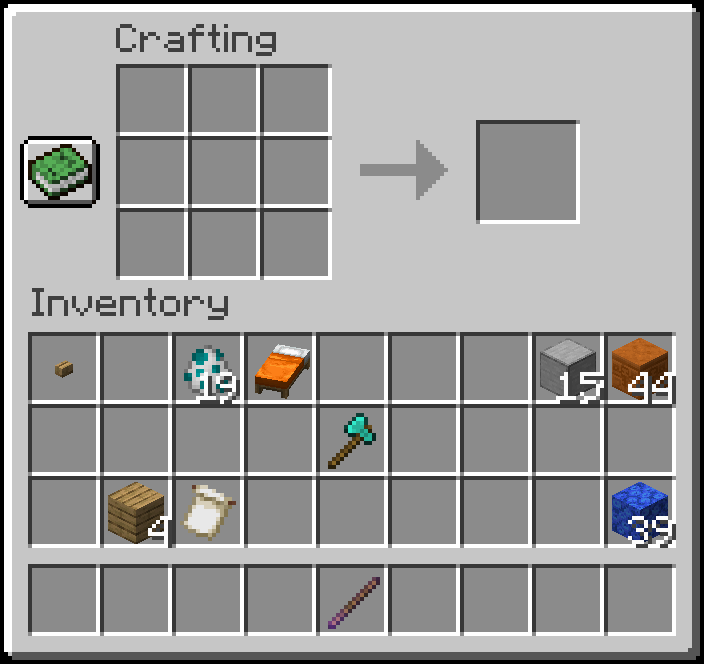}\\
                \vspace{0.5em}
            \end{minipage}
        \end{minipage}
    \end{tcolorbox}
    \caption{Failure example from Memory-Only with an \executable teacher. The agent retrieves a memory (cache hit) that contained a plan to craft an oak boat, however it is not directly applicable for the current task. This leads to an example of what we call \textbf{eager crafting}, where the agent crafts an object that is not only unnecessary but also whose crafting prevents ever reaching the target.}   \label{fig:qualitative_mem_executable_failure_VAL0356}
\end{figure*}

\begin{figure*}[h!]
    \begin{tcolorbox}[title={Success: VAL0452 (Target: acacia\_pressure\_plate)}, colbacktitle=greenPP, unbreakable]
        \small
        \begin{minipage}{\textwidth}
            \begin{minipage}{0.65\textwidth}
                \textbf{Dialogue:}\\
                \textcolor{black}{\textbf{User:}} Craft an item of type: acacia\_pressure\_plate\\
                inventory:\\
                - carved\_pumpkin I2 quantity 19\\
                - music\_disc\_pigstep I3 quanti...\\
                \vspace{0.5em}
                \textcolor{yellowPP}{\textbf{read\_memory(\{"recipe": "acacia\_pressure\_plate"\})}}\\
                \vspace{0.5em}
                \textcolor{redPP}{\textbf{Cache Miss}}\\
                \vspace{0.5em}
                \textcolor{greenPP}{\textbf{Question}} How do I craft an acacia\_pressure\_plate using the items in my inventory, specifically with the available acacia\_planks?\\
                \vspace{0.5em}
                \textcolor{greenPP}{\textbf{Answer:}} To craft an acacia\_pressure\_plate, first arrange two acacia\_planks in a 1x2 shape in the top row of the crafting grid, with one acacia\_plank in the top left and the other in the top middle. Then, the acacia\_pressure\_plate will be created and can be moved to a free inventory slot.\\
                \vspace{0.5em}
                \textcolor{greenPP}{\textbf{Parsed Answer:}} RECIPE: acacia\_pressure\_plate\\
                REQUIREMENTS: \\
                - 2 acacia\_planks\\
                PROCEDURE: \\
                1. Arrange 2 acacia\_planks in a 1x2 shape in the top row of the crafting grid.\\
                2. Place one acacia\_plank in the top left (A1) and the other in the top middle (A2).\\
                3. The acacia\_pressure\_plate will be created in the output slot.\\
                4. Move the acacia\_pressure\_plate from the output slot to a free inventory slot.\\
                RELATED IT...\\
                \vspace{0.5em}
                \textcolor{yellowPP}{\textbf{move(\{"slot\_from": "I32", "slot\_to": "A1", "quantity": 1\})}}\\
                \vspace{0.5em}
                \textcolor{black}{\textbf{User:}} Craft an item of type: acacia\_pressure\_plate\\
                inventory:\\
                - acacia\_button 0 quantity 1\\
                - acacia\_planks A1 quantity 1\\
                - ...\\
                \vspace{0.5em}
                \textcolor{yellowPP}{\textbf{move(\{"slot\_from": "I32", "slot\_to": "A2", "quantity": 1\})}}\\
                \vspace{0.5em}
                \textcolor{black}{\textbf{User:}} Craft an item of type: acacia\_pressure\_plate\\
                inventory:\\
                - acacia\_pressure\_plate 0 quantity 1\\
                - acacia\_planks A1 quanti...\\
                \vspace{0.5em}
                \textcolor{yellowPP}{\textbf{think(\{"thought": "I need to move the acacia\_pressure\_plate from the output slot to a free inventory slot."\})}}\\
                \vspace{0.5em}
                \textcolor{yellowPP}{\textbf{move(\{"slot\_from": "0", "slot\_to": "I1", "quantity": 1\})}}\\
                \vspace{0.5em}
                \textcolor{greenPP}{\checkmark~\textbf{Crafting succeeded!}}
            \end{minipage}
            \hfill
            \begin{minipage}{0.32\textwidth}
                \textbf{Environment Frames:}\\
                \textbf{Frame 1:}\\
                \includegraphics[width=\textwidth]{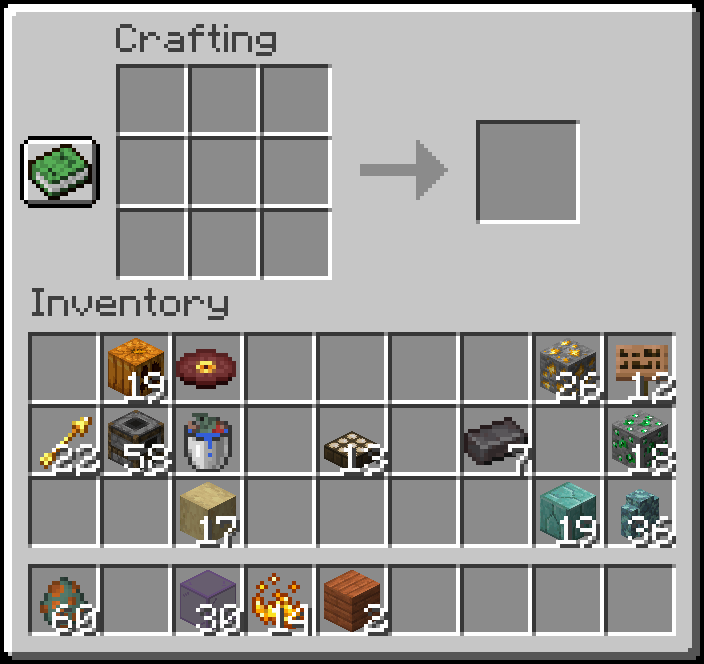}\\
                \vspace{0.5em}
                \textbf{Frame 2:}\\
                \includegraphics[width=\textwidth]{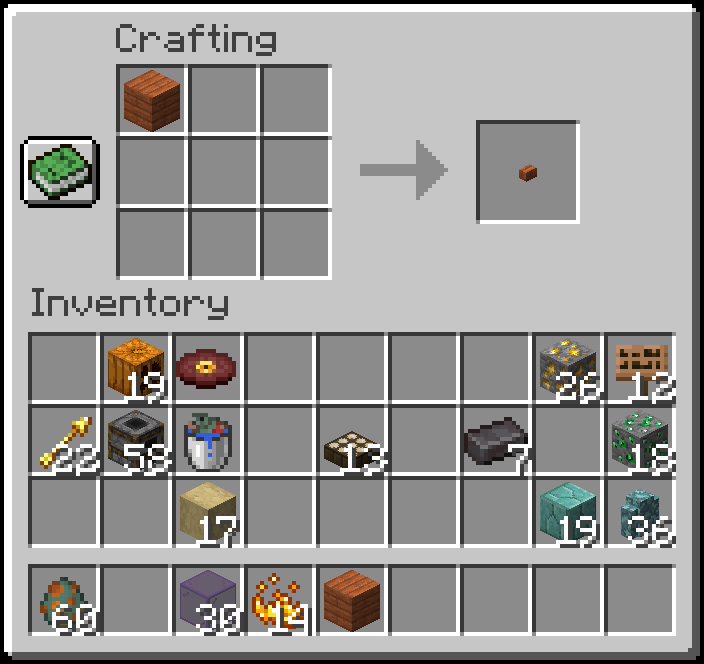}\\
                \vspace{0.5em}
                \textbf{Frame 3:}\\
                \includegraphics[width=\textwidth]{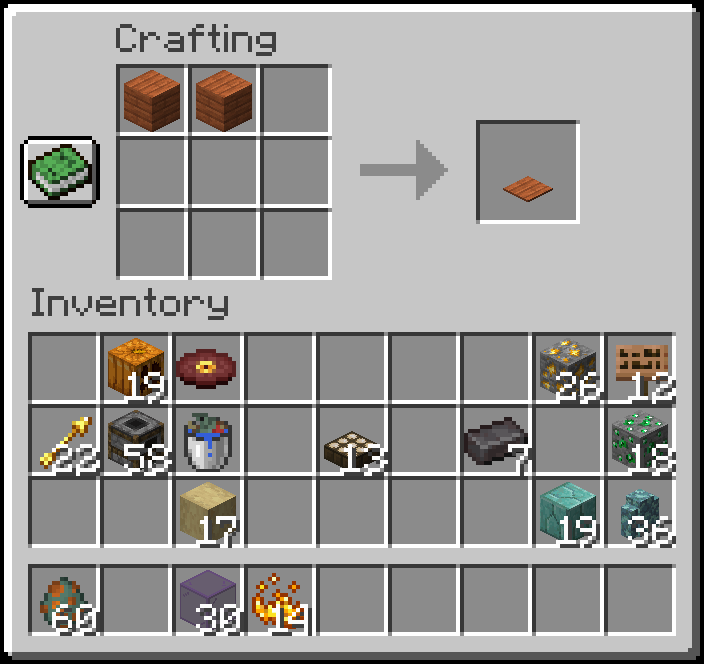}\\
                \vspace{0.5em}
            \end{minipage}
        \end{minipage}

    \end{tcolorbox}
    \caption{Success example from $How^2$ with a \nonexecutable teacher. The agent reads the memory for an acacia pressure plate, since there are no relevant memories, it asks a how-to question to the teacher. The teacher answers in the generic recipe that is entirely ungrounded in the Plancraft environment. The parse step, ground the answer to actual slots. The agent then uses the parsed memory to successfully craft the target item. }
    \label{fig:qualitative_mem_non_executable_parse_relevance_success_VAL0452}
\end{figure*}

\begin{figure*}[h!]
    \begin{tcolorbox}[title={Failure: VAL0288 (Target: brown\_banner)}, colbacktitle=redPP, unbreakable]
        \small
        \begin{minipage}{\textwidth}
            \begin{minipage}{0.65\textwidth}
                \textbf{Dialogue:}\\
                \textcolor{black}{\textbf{User:}} Craft an item of type: brown\_banner\\
                inventory:\\
                - brown\_wool I7 quantity 6\\
                - terracotta I9 quantity 19\\
                - stick I14 qua...\\
                \vspace{0.5em}
                \textcolor{yellowPP}{\textbf{read\_memory(\{"recipe": "brown\_banner"\})}}\\
                \vspace{0.5em}
                \textcolor{redPP}{\textbf{Cache Miss}}\\
                \vspace{0.5em}
                \textcolor{greenPP}{\textbf{Question}} How do I craft a brown\_banner using the items in my inventory?\\
                \vspace{0.5em}
                \textcolor{greenPP}{\textbf{Answer:}} To craft a brown\_banner, first arrange six brown\_wool items in the top two rows of the crafting grid in a 3x2 pattern, filling all the top and middle row slots, and then place one stick in the bottom middle slot of the crafting grid. This will produce a brown\_banner, which can then be moved to a free inventory slot.\\
                \vspace{0.5em}
                \textcolor{greenPP}{\textbf{Parsed Answer:}} RECIPE: brown\_banner\\
                REQUIREMENTS: \\
                - 6 brown\_wool\\
                - 1 stick\\
                PROCEDURE: \\
                1. Arrange 6 brown\_wool items in a 3x2 pattern, filling all slots in the top and middle rows of the crafting grid.\\
                2. Place 1 stick in the bottom middle slot (C2) of the crafting grid.\\
                3. Move the resulting brown\_banner from the output slot to a free inventory slot.\\
                RELATED ITEMS: ['brown\_wool', 'stick']\\
                \vspace{0.5em}
                \textcolor{yellowPP}{\textbf{move(\{"slot\_from": "I7", "slot\_to": "A1", "quantity": 1\})}}\\
                \vspace{0.5em}
                \textcolor{black}{\textbf{User:}} Craft an item of type: brown\_banner\\
                inventory:\\
                - brown\_wool A1 quantity 1\\
                - brown\_wool I7 quantity 5\\
                - terracotta I9 ...\\
                \vspace{0.5em}
                \textcolor{yellowPP}{\textbf{move(\{"slot\_from": "I7", "slot\_to": "A2", "quantity": 1\})}}\\
                \vspace{0.5em}
                \textcolor{black}{\textbf{User:}} Craft an item of type: brown\_banner\\
                inventory:\\
                - brown\_carpet 0 quantity 3\\
                - brown\_wool A1 quantity 1\\
                - brown\_wool A2...\\
                \vspace{0.5em}
                \textcolor{yellowPP}{\textbf{move(\{"slot\_from": "0", "slot\_to": "I35", "quantity": 3\})}}\\
                \vspace{0.5em}
                \textcolor{redPP}{\texttimes~\textbf{Crafting failed!}}
            \end{minipage}
            \hfill
            \begin{minipage}{0.32\textwidth}
                \textbf{Environment Frames:}\\
                \textbf{Frame 1:}\\
                \includegraphics[width=\textwidth]{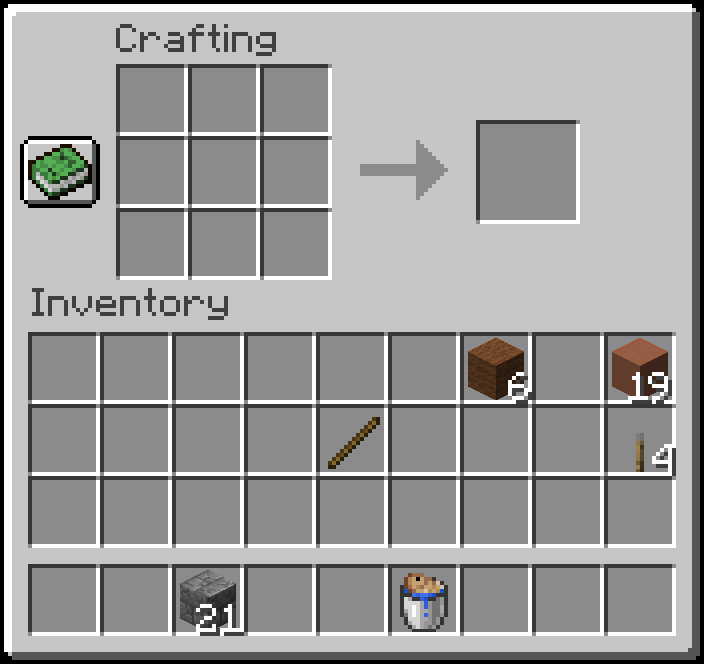}\\
                \vspace{0.5em}
                \textbf{Frame 2:}\\
                \includegraphics[width=\textwidth]{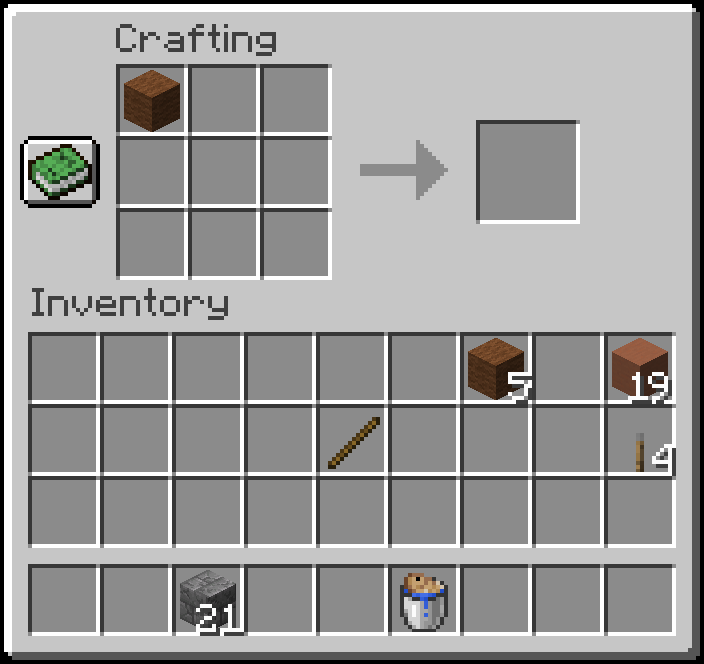}\\
                \vspace{0.5em}
                \textbf{Frame 3:}\\
                \includegraphics[width=\textwidth]{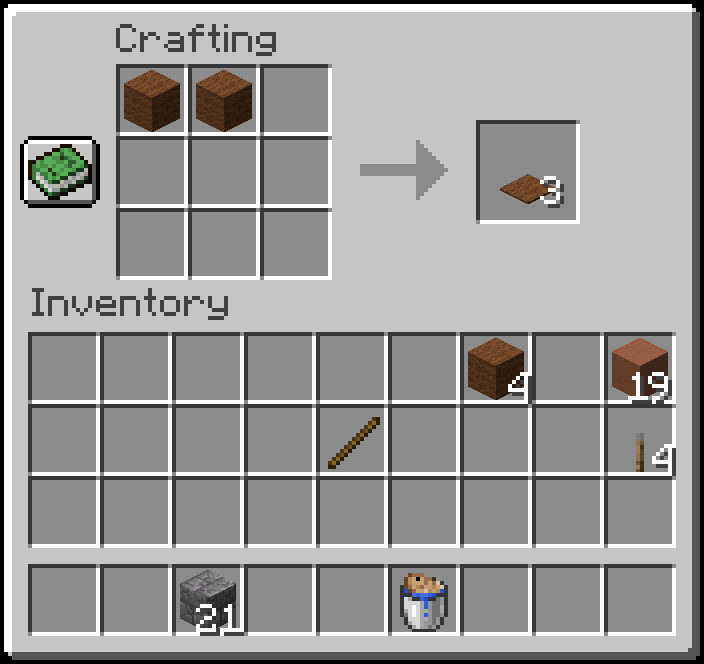}\\
                \vspace{0.5em}
            \end{minipage}
        \end{minipage}

    \end{tcolorbox}
    \caption{Failure example from $How^2$ with a \nonexecutable teacher. The teacher provides an ungrounded explanation of the pattern, which the parsed step translates into a structured memory. In this case, the parse step fails to ground the 3x2 pattern to the 6 relevant crafting slots. As the agent starts filling the crafting grid and following the instructions, the brown carpet is added to the output slot as its pattern is a subset of the brown banner. This leads to another example of \textbf{eager crafting}, where the agent crafts an item present in the output slot even though it is suboptimal and leads to failure.}
    \label{fig:qualitative_mem_non_executable_parse_relevance_failure_VAL0288}
\end{figure*}

In Figures~\ref{fig:qualitative_ask_subgoal_partially_executable_success_VAL0531}, ~\ref{fig:qualitative_ask_subgoal_partially_executable_failure_VAL0540}, ~\ref{fig:qualitative_mem_executable_success_VAL0336},
~\ref{fig:qualitative_mem_executable_failure_VAL0356},
~\ref{fig:qualitative_mem_non_executable_parse_relevance_success_VAL0452} and \ref{fig:qualitative_mem_non_executable_parse_relevance_failure_VAL0288} we show examples of successful and failed trajectories from runs with different setups and teachers.
Tool calls are highlighted in \textcolor{yellowPP}{yellow}, memory reads in \textcolor{greenPP}{green}, and user inputs in \textcolor{black}{black}.
User observations are highlighted in gray boxes with corresponding environment frames shown on the right.

The figures illustrate both successful and failed crafting attempts across different experimental setups.
Successful trajectories (Figures~\ref{fig:qualitative_ask_subgoal_partially_executable_success_VAL0531}, \ref{fig:qualitative_mem_executable_success_VAL0336}, and \ref{fig:qualitative_mem_non_executable_parse_relevance_success_VAL0452}) show the agent effectively using teacher-provided plans, whether they are partially executable, fully executable, or require parsing from natural language.
In contrast, failure cases highlight specific challenges.
These include the agent failing to consult its memory and prematurely declaring a task impossible (Figure~\ref{fig:qualitative_ask_subgoal_partially_executable_failure_VAL0540}), or instances of \textit{eager crafting} where the agent crafts an item that renders the target unreachable (due to lack of resources).
Both Figure~\ref{fig:qualitative_mem_executable_failure_VAL0356} and Figure~\ref{fig:qualitative_mem_non_executable_parse_relevance_failure_VAL0288} show the agent incorrectly craft the wrong item.
In the first case, this is most likely due to the agent following an inapplicable cached plan.

